\documentclass{article}
\usepackage{amsmath, amsthm, amssymb}
\usepackage[mathscr]{eucal}
\usepackage{graphicx}
\usepackage{verbatim}
\usepackage{caption}
\usepackage{subcaption}
\usepackage{accents}
\usepackage{comment}
\usepackage{bm}
\usepackage[margin=1.0in]{geometry}
\usepackage{natbib}
\usepackage{url}
\usepackage{adjustbox}
\usepackage{multicol}
\usepackage{booktabs}
\usepackage{pdflscape}
\usepackage[breaklinks, hidelinks]{hyperref} % this disables the boxes around links
\setcitestyle{authoryear}
\usepackage[disable,colorinlistoftodos,prependcaption,textsize=small,backgroundcolor=yellow,linecolor=lightgray,bordercolor=lightgray]{todonotes}

\DeclareMathOperator*{\argmin}{arg\,min}
\DeclareMathOperator*{\argmax}{arg\,max}
\newcommand{\vect}[1]{\ensuremath{\boldsymbol{#1}}}

\newcommand{\midw}{0.7\textwidth}
\newcommand{\fullw}{\textwidth}
\newcommand*{\transpose}{{\mathpalette\@transpose{}}}

\let\oldfootnote\footnote
\renewcommand{\footnote}{\unskip\oldfootnote}% Remove any skips inserted before \footnote

\title{Counterfactual Inference for Consumer Choice Across Many Product Categories}
\author{Rob Donnelly, Francisco J. R. Ruiz, David Blei, and Susan Athey\thanks{
    We follow the machine learning tradition in author ordering, with Rob Donnelly having the greatest contribution.
    Athey: Stanford University, 655 Knight Way, Stanford, CA 94305, athey@stanford.edu.
    Blei: Columbia University, Department of Computer Science, New York, NY, 10027, david.blei@columbia.edu.
    Donnelly: Instacart, 50 Beale St \#600, San Francisco, CA 94105, rndonnelly@gmail.com.
    Ruiz: University of Cambridge, Department of Engineering, Cambridge CB2 1PZ,
    franrruiz@deepmind.com
}}

\begin{document}

\maketitle

\begin{abstract}
This paper proposes a method for estimating consumer preferences among discrete choices, where the
consumer chooses at most one product in a category, but selects from multiple categories in parallel.
The consumer's utility is additive in the different categories.  Her preferences about product
attributes as well as her price sensitivity vary across products and may be correlated across products.
We build on techniques from the machine learning literature on probabilistic models of matrix
factorization, extending the methods to account for time-varying product attributes and products
going out-of-stock.
We evaluate the performance of the model using held-out data from weeks with price changes or out of
stock products.
We show that our model improves over traditional modeling approaches that consider each category in
isolation.
One source of the improvement is the ability of the model to accurately estimate heterogeneity in
preferences (by pooling information across categories); another source of improvement is its ability
to estimate the preferences of consumers who have rarely or never made a purchase in a given
category in the training data.
Using held-out data, we show that our model can accurately distinguish which
consumers are most price sensitive to a given product.
We consider counterfactuals such as personally targeted price discounts, showing that using a richer
model such as the one we propose substantially increases the benefits of personalization in
discounts.
\end{abstract}


\section{Introduction}
Estimating consumer preferences among discrete choices has a long history in economics and
marketing.
\citeauthor{Domencich1975}'s \citeyearpar{Domencich1975} pioneering analysis of transportation
choice articulated the benefits of using choice data to estimate latent parameters of user utility
functions \citep[see also][]{Hausman1978}: once estimated, a model of user utility can be used to
analyze counterfactual scenarios, such as the impact of a change in price or of the introduction of
an existing product to a new market.
McFadden (1974) also highlighted strong assumptions implicit in using off-the-shelf multinomial
choice models to estimate preferences and introduced variants such as the nested logit that relaxed
some of the strong assumptions (including ``independent of irrelevant alternatives'').

Analysts have applied the discrete choice framework to a variety of different types of data sets,
including aggregate, market-level data (see, e.g., \citet{Berry1995}, \citet{Nevo2001}, and
\citet{Petrin2002})\footnotemark{}, as well as data from individual choices for a cross-section of
individuals.
\footnotetext{This literature grapples with the challenge that to the extent prices vary across
  markets, the prices are often set in response to the market conditions in those markets. In
  addition, to the extent that products have quality characteristics that are unobserved to the
  econometrician, these unobserved quality characteristics may be correlated with the price.}
In this paper, we focus on models designed for a particularly rich type of data, consumer panel
data, where the same consumer is observed making choices over a period of time.
Supermarket scanner data is a classic example of this type of data, but e-commerce firms also
collect panel data and use it to optimize their offerings and prices.
Scanner data enables the analyst to enrich the analysis in a variety of ways, for example to account
for dynamics (see \citet{Keane2013}) for a survey.
The vast majority of the literature based on individual choice data focuses on one
category\footnotemark{} at the time, e.g.\ \citet{Ackerberg2001,Ackerberg2003a} analyzes yogurt,
\citet{Erdem2003} ketchup, \citet{Dube2004} soft drinks, and \citet{Hendel2006a} detergent.
\footnotetext{Throughout the paper we use category to refer to disjoint sets of products, such that
  products that are within the same category are partial substitutes.}
Often these analyses focus on the impact of marketing interventions, such as advertising campaigns,
coupons, or promotions, such as in  the classic work by \citet{Rossi1996}.

In this paper, we analyze the demand for a large number of categories in parallel.
This approach has a number of advantages.
First, there is the potential for large efficiency gains in pooling information across categories
if the consumer's preferences are related across categories.
For example, the consumer's sensitivity to price may be related across categories, and there may be
attributes of products that are common across categories (such as being organic,
convenient, healthy, or spicy).
These efficiency gains are likely to be particularly pronounced for less commonly purchased items.
Even among the top 100 categories in a supermarket, the baseline probability of purchasing an item
in the category is very low on any particular trip, and there are thousands of
categories in a typical store.%
\footnote{In the sample we use in our empirical exercise, among the top 123 categories the average
  category is only purchased on 3.7\% of shopping trips. Only milk, lunch bread, and tomatoes are
  purchased on more than 15\% of trips. Things are even sparser at the individual UPC level. The
  average purchase rate is 0.36\% and only one, avocados, is purchased in more than 2\% of the trips
  in our Tuesday-Wednesday sample.}
For e-commerce, there may be millions of products, most of which are rarely or never purchased by
any particular consumer.
But by pooling data across categories, it is possible to make personalized predictions about
purchasing, even for categories in which the consumer has not purchased in the past.

A second advantage of analyzing many categories at once is that from the perspective of marketing,
it is crucial for retailers to understand their consumers in terms of what drives their overall
demand at the store, not just for individual products.
For example, there may be products that are very important to high-volume shoppers, but where they
are price-elastic; avoiding stock-outs on those products and offering competitive prices may be very
important in store-to-store competition.
Although this paper does not offer a complete model of consumers' choice across stores, we view the
demand model we introduce as an important building block for such a model.

Our model makes use of recent advances in machine learning and scalable Bayesian modeling to
generate a model of consumer demand.
Our approach learns a concise representation of consumer preferences across multiple product
categories that allows for rich (latent, i.e.\ unobservable) heterogeneity in products as well as
preferences across consumers.
Our model assumes that consumers select a single item from a given category (a strong form of
substitution, where in the empirical analysis we drop categories that have large violations of this
assumption), and further assumes that purchases are independent across categories (thus ignoring
budget constraints, which we argue are less likely to bind at the level of an individual shopping
trip).
From a machine learning perspective, we extend matrix factorization techniques developed by
\citet{Gopalan2013} to focus on the case of shopping, which requires incorporating time-varying
prices and demand shifts as well as an appropriate functional form.
We introduce ``sessions,'' where prices and the availability of products are constant within a
session; but these elements may change across sessions.
It is common in stores for products to go in and out of stock, or to be promoted in various ways;
accounting for these factors is helpful in allowing the model to estimate the parameters that are
most useful for counterfactual inference.
Finally, relative to the machine learning approach, we tune our model hyperparameters on the basis of
performance on counterfactual estimates, and we show that this makes a difference relative to
focusing on the typical machine learning objective, prediction quality. We are able to do this because
our data contains a large number of distinct price changes and examples of products going out of stock,
and thus we can hold out data related to some of these changes and evaluate performance of the
model in predicting the impact of those changes.

The primitives of our model include the latent characteristics of products (a vector, whose
dimension is tuned in the process of estimation on the basis of goodness of fit), as well as each
consumer's latent preferences for each dimension.
These latent characteristics and preferences are constant over time.
In addition, we do not assume that the consumer's price sensitivity is constant across products;
instead, each product has a vector of latent characteristics that relate to consumers' price
sensitivity toward the product, and each consumer's price sensitivity is the inner product of a
consumer-specific latent vector and the product's latent characteristics that relate to price
sensitivity.
Thus, both the mean utility and the price sensitivity are allowed to flexibly vary across
consumers and products.%
\footnote{We also ran alternative specifications with the per consumer price
  coefficients restricted to be the same across all products, however this lead
  to a substantial reduction along both the predictive and counterfactual fit
  measures of performance.
}
In addition, the model includes controls for week-specific demand for product categories.
We use a Bayesian approach, so our model produces a posterior distribution over each latent factor.

We apply our model to data from a single supermarket over a period of 23 months, where we observe
the same consumers shopping over time.
The data originate from shopper loyalty cards.
Unlike many panels collected by third parties, the data is available at the level of the trip rather
than aggregated to the weekly level, and we see shopping at a high enough frequency to identify the
timing of price changes and stock-outs.
In particular, we observe many weeks where prices change at midnight on Tuesday night; and
otherwise, behavior is very similar between Tuesday and Wednesday.
This allows us to identify the effects of price changes and be able to make counterfactual
predictions.
We conduct a variety of tests that assess our identification strategy, and in a departure from the
machine learning literature on which we build, we evaluate model fit on the basis of the model's
ability to predict how behavior changes when prices change.
We compare our model to a variety of commonly-used category-by-category models, including nested
logit and mixed logit, showing that our model performs better both in terms of overall ability to
fit on a representative test set, but also in terms of the model's ability to predict responses to
price changes.
We examine both own-price and cross-price effects.
We also examine whether the heterogeneity incorporated in our model is spurious or predictive by
showing that our model tends to produce more heterogeneity across groups in terms of own-price and
cross-price elasticities, and using held-out data, we show that this heterogeneity predicts
heterogeneity in consumer response to price changes.
We also show that our model has key advantages in terms of being able to predict the behavior of
consumers who have rarely or never purchased in the training data.

\section{Related Work}
\label{sec:relatedwork}

In the traditional discrete choice literature, it has become common to include many latent variables
describing user preferences; for example, in a mixed logit model, it is common to include a
user-item random effect, as well as individual-specific preference parameters for price and other
observed item attributes \citep[e.g.][]{Berry2004a, Train2009}.
However, it is less common to model latent item attributes, other than perhaps a single dimension
(quality).
There are, however, several lines of work that estimate richer models that incorporate latent user
characteristics, exploiting panel data.%
\footnote{In data from a single cross-section of consumers, \citet{Athey2007a} show that only a
  single latent variable can be identified (or two if utility is restricted to be monotone in each)
  without functional form restrictions, arguing that panel data is critical to uncover common latent
  characteristics of products.}

An early example is the ``market mapping'' literature, where each product is described as a vector
of latent attributes.
A market map can be used for a variety of exercises; for example, one can consider the entry of a
new product into a position in the product space and forecast which consumers are likely to buy it.
Empirical applications have also typically focused on a single category, such as laundry detergent
(\citet{Elrod1995} and \citet{Chintagunta1994}).%
\footnote{\citeauthor{Elrod1995} use a factor analytic probit model with normally distributed
  preferences, whereas \citeauthor{Chintagunta1994} uses a logit model with discrete segments of
  consumer types.\citet{Elrod1988, Elrod1988a} use logit models to estimate up to two latent
  attributes and the distribution of consumer preferences. The former study uses a linear utility
  specification, and the latter uses an ideal-point model.}
Outside of shopping, there are several other social science applications making use of panel data to
estimate latent item attributes and individual preferences.
\citet{Goettler2001} study television viewing for a panel of users, and attempt to estimate latent
attributes of television shows based on this panel.
Another application area is political science, where panel data on legislators' voting decisions is
used to uncover their preferences and the latent characteristics of legislation.
\citet{Poole1985} use a transformed logit model to estimate both the locations of legislators’ ideal
points and the locations of legislative bills in a unidimensional attribute space.

There has been some progress on estimating multiple-discrete choice models in which consumers choose
more than one of a single item \citep[e.g.][]{Hendel1999,Kim2002,Dube2004}, however a substantial
portion of the literature continues to focus on categories in which the unit-demand assumption
plausibly holds.

Despite the extensive literature making use of consumer panel data, very little literature in
economics and marketing attempts to consider multiple categories simultaneously.
A few papers study demand for bundles of products, where the products may be substitutes or
complements, and where the models attempt to estimate the nature of interaction effects.
These models are limited by the curse of dimensionality and generally have difficulty incorporating
more than two or three categories (e.g.\ \citet{Athey1998}; see \citet{Berry2014} for a review).
The only paper we are aware of that estimates interaction effects across many categories is
\citet{Ruiz2020} which uses a similar approach to this paper, but focuses on estimating
interactions rather than exploiting available information about the category structure.
We discuss this in more detail below.

Our model focuses on sharing information about consumer preferences for item attributes across
categories where consumer preferences are additively separable across categories.
Our model differs from the past literature in social sciences in the techniques used and in the
scale and complexity of the model.
In order to flexibly estimate consumer heterogeneity across multiple product categories, this paper
builds on the Bayesian Hierarchical Poisson Factorization (HPF) model proposed in \citet{Gopalan2013}.
The HPF model predicts the preferences each user (decision maker) has for each item (product) based
on a sum of the product of a latent vector of item characteristics and a latent vector of consumer
preferences for each of those item characteristics.%
\footnote{This approach has similarities to the econometrics literature on ``interactive fixed
  effects models'' although that literature has focused primarily on decomposing common trends
  across individuals over time rather than identifying common preferences for products across
  individuals \citep{Moon2015, Moon2014, Bai2009}}

\citet{Gopalan2013} demonstrate that the HPF model can make accurate predictions\footnotemark{} across
a wide variety of contexts, including Netflix movies, New York Times articles, and scientific
articles in a researcher's Mendeley account.
\footnotetext{As is standard in the machine learning literature, the accuracy is estimated on a ``held
  out'' or ``test'' data set that is not used during the training of the model. This is a more
  accurate way to evaluate how well a model will be able to make predictions on new data that has
  not yet been observed.}
Despite the reputation that Bayesian methods have for being slow computationally, this model is
scalable across large data set sizes\footnotemark{} due to its use of mean-field variational inference
to approximate the computationally intractable exact posterior.
\footnotetext{For example, when trained on Netflix data with 480,000 users, 17,700 movies, and 100
  million observations, they report the model took 13 hours to converge on a single CPU.}

The HPF model is related to the extensive recommender systems literature that uses
matrix-factorization-based techniques to predict what items (movies, links, articles, search
results, etc.) a user will enjoy based on their previous choice behavior
\citep{Koren2009,Bobadilla2013}.
A core insight of this literature is that it is often very effective to predict a user's interests
based on the preferences of other users who have similar tastes.
These approaches try to find a lower dimensional approximation of the full matrix of user and item
preferences.%
\footnote{The matrix has one row for each user and one column for each item. The $i,j$ entry
  corresponds to how much user $i$ ``likes'' item $j$. We often only observe some of the entries of
  this large matrix, and would like to make predictions for the unobserved entries. e.g.\ predict how
  much a user will like a movie that they haven't watched or rated yet.}
The resulting factorization often is able to make accurate predictions and can also provide an
interpretable representation of the user preferences in the data.

In our empirical exercise, using data from a supermarket loyalty program, we show that simultaneously
modeling consumers' decisions across multiple product categories helps improve our ability to
characterize individual level preferences relative to estimating preferences in each category
independently.
This has similarities to the growing area of transfer learning in machine learning
\citep{Pan2010,Oquab2014} in which training a model on one domain (e.g.\ one for which large amounts
of data are available) can help improve the model's ability to make predictions in a different
domain (potentially one for which less data is available).
This insight may have applications in other economics and marketing contexts, for example, data on
consumers' purchasing decisions in one domain in which purchases are frequent (grocery stores) may
be able to improve our ability to estimate consumer demand in a seemingly unrelated domain in which
purchases are much less frequent (cars).

Some researchers have begun using latent-factorization-based approaches in the
context of customer purchasing habits in grocery and retail.
\citet{Jacobs2016} extend a widely-used approach known as latent Dirichlet allocation (LDA) towards the
task of predicting consumer purchases in online markets with large product assortments.
More recently, \citet{Jacobs2021} further extended their approach to model a
customer's purchasing intents across multiple shopping trips and allow a
customer to have multiple distinct motivations within the latent factorization.
\citet{Wan2017} propose an alternative model for consumer grocery demand
estimation which uses a latent factorization model with three stages.
In the first stage, the consumer makes a binary choice for each category about whether
or not to purchase something from the category.
Next, the consumer makes a multinomial choice of which item from the category to
purchase.
Finally, the consumer makes a choice of how many of this item to purchase, drawn
from a Poisson distribution.
Latent factorization is carried out independently for each stage as regularized
maximum likelihood estimation.
\citet{Ruiz2020} use a model closely related to ours and the same grocery
purchase data with a focus on heuristically identifying which products are
likely to be substitutes or complements to each other.

In comparison to these papers, this paper differs in its focus on designing a
model able to accurately make counterfactual predictions about how customers will
respond if prices were changed or the set of available products were different,
for example, due to a product being out-of-stock.
This goal motivates our introduction of two-stage nesting structure inspired by
nested logistic regression, which allows the model to capture more realistic
patterns of substitution between products.
This paper also systematically evaluates the assumptions required to identify price elasticities,
and conducts a comparison of alternative category-by-category demand models.
In Section~\ref{sec:ModelFit}, we discuss in more detail the
variety of approaches we use to evaluate the models. In particular, we show
several ways to go beyond the focus on predictive fit in held-out
data that is common in the machine learning literature.
These approaches include evaluating our predictions across large numbers of what
we call mini quasi-experiments, subsets of the held-out data that
approximate the variation we would like to make counterfactual
predictions about.
In our application, we focus on pairs of sequential days in which either a
product changes price or the set of available products changes due to a stock
out, and evaluate the ability of models to predict the changes in
individual-level and aggregate demand.
We also evaluate the ability of each model to capture ``intuitive priors'',
beliefs that we expect to hold in a well-functioning model.
For example, we look at whether the models predict higher cross-price
elasticities between products that are more similar along characteristics
that were not used while training the models.
Finally, we contrast two approaches for evaluating the business impact of
marketing decisions that could be powered by the model predictions.
The first approach, which is common in the economics and marketing literature,
is to use a model to predict impact of an alternative treatment policy, for example an allocation of
coupons to a particular subset of customers.
The impact of the new policy is estimated by
treating the model's fitted data generating process as ground truth.
This approach can be useful for comparing the performance of different treatment
policies, as shown in the seminal paper on the value of purchase history data in
target marketing by \citet{Rossi1996}.
However treating a model's predictions as ground truth can overestimate
the true impact, especially if one does not properly account for the uncertainty
in the fitted data generating process.
This over-confidence can be especially significant when using models that allow
rich individual-level heterogeneity such as ours.
An alternative approach is to evaluate and compare different pricing policies
directly in the held-out data in a model-free way inspired by the policy
learning literature \citep{Dudik2011,Zhou2018}.
We demonstrate this approach by selecting the two most common prices for each
product in the data and using the model to assign each user to the price that is
predicted to lead to higher profits.
We can then evaluate in the held-out shopping trips whether we do in fact earn
higher profits per trip when customers shop on days with their assigned price.

\section{The Model}

\subsection{Random Utility Models and Independence of Irrelevant Alternatives}
\label{sec:IIA}

In this section, we introduce the canonical random utility model (RUM).
Consider shopper $i$ on a shopping trip at time $t$.
In each product category, $c = 1 \ldots C$ (e.g.\ bananas, laundry detergent, yogurt), there are $j = 1
\ldots J_c$ products to choose between.
Within each category, the shopper has unit demand and will purchase at most one item.
To simplify the model, we assume that the product categories are disjoint and that there is no
substitution or complementarity between products in separate categories.%
\footnote{See \cite{Ruiz2020} for a related paper that uses a heuristic approach to identify
  potential substitutes and complements automatically based on patterns of co-purchase.}
The shopper purchases the item that provides her with the highest utility among the options in the
category.
\begin{align*} \label{eq:RUM}
   U_{ijt} &= u_{ijt} + \epsilon_{ijt} \\
   y_{ijt} = 1 \text{ if } j &= \argmax_{k \in 1 \ldots J_c} U_{ikt}
\end{align*}

If we assume that the $\epsilon_{ijt}$ are drawn i.i.d.\ from an extreme value type 1 distribution\footnotemark{}, then
\footnotetext{Also known as the Gumbel distribution.}
\begin{equation} \label{eq:Logit}
  P \left( y_{ijt} = 1 \right)
  = \frac{\exp(u_{ijt})}{ \sum_{j' = 1}^{J_c} \exp(u_{ij't})}
\end{equation}

The ratio of purchase probabilities between any two items $j$ and $j'$ in the same category depend
only on the ratio of their $u$ values.
\begin{equation}
  \frac{P(y_{ijt}=1)}{P(y_{ij't}=1)}
  = \frac{\exp(u_{ijt})}{\exp(u_{ij't})}
\end{equation}

Similarly if we consider some subset of the products in a category $S_c \subset J_c$, then
conditional on the purchase of an item from the subset, the relative purchase probability for item
$j \in S_c$ is given by

\begin{equation}
  P \left(
    y_{ijt} = 1
    \mid \sum_{j' \in S_c} y_{ij't} = 1, \sum_{j' \notin S_c} y_{ij't} = 0
  \right)
  = \frac{\exp(u_{ijt})}{\sum_{j' = 1}^{S_c} \exp(u_{ij't})}\label{eq:IIA}
\end{equation}

This property is known as ``independence of irrelevant alternatives'' (IIA).
IIA imposes strong constraints on the patterns of substitution
between products due to the assumption of the independence of the error terms $\epsilon_{ijt}$.
Suppose that product $j$ became unavailable or its attractiveness to the
consumer decreases due to an increase in price.
Under the assumption of IIA, the resulting customer choices will be reallocated
proportionally to their initial levels; there can be no differential level
of substitutability between products.
How problematic this is in practice depends on what factors are included in the
model of the $u_{ijt}$ terms.
A model with no heterogeneity in preferences across users would be unable to
capture the intuitive result that, when one product becomes unavailable,
the other products most similar to it to gain a
disproportionate fraction of the displaced purchases.
Similarly, when predicting the effect of changing one product's price on the
demand for other products in the same category (the ``cross-price elasticity''),
the predicted effect will not depend on the similarity of the products.

This undesirable implication can be partially mitigated by extending the model
to allow customer preferences to vary across the population, even if each individual's
demand is still assumed to satisfy IIA.
For example, suppose that some customers like spicy salsa and other customers
like mild salsa.
Within each group, customer demand satisfies IIA and they proportionally
substitute between the salsas that match their tastes. However, when we aggregate
across the full population, we get the intuitive result that when one brand of
spicy salsa goes on sale, it steals more market share from the other spicy
salsas than from the mild salsas.%
\footnote{\citet{McFadden1974} provides a well known
  thought experiment illustrating this effect in the context of commuters
  choosing between driving a car, riding a red bus, or riding a blue bus.
  \citet{Steenburgh2013} provides further details on the degree to which
  allowing heterogeneity in preferences reduces (but does not eliminate) the
  problems of the homogeneous logit model.}
However, even with rich heterogeneity in customer preferences, assuming IIA may
still be problematic if there are unobserved factors that cause correlations in
an individual consumer's choice probabilities across time.
For example, the weather, the contents of the household's cupboards, or
the shopper's mood, might simultaneously affect the utilities $U_{ijt}$ of
multiple products at the same time.%
\footnote{In the context of the model, this would manifest as a correlation in
  the error terms $\epsilon_{ijt}$ across products, which the model assumes to be
  independent and identically distributed.}
Allowing customers to have different purchasing intentions on different shopping
trips is one way to potentially address this limitation \citep[e.g.][]{Jacobs2021}.

\subsection{Estimation of the RUM using Nested Hierarchical Factorization Model}

\subsubsection{Nested Factorization Model}
\label{sec:NF}

The Nested Factorization model builds on the Hierarchical Poisson Factorization (HPF) model proposed
by \cite{Gopalan2013}, adding a number of additional features important for capturing shopping
behavior.
It also extends the Time Travel Factorization Model (TTFM) introduced in \cite{Athey2018}.
The TTFM model predicts the choice of where to go to lunch, treating all
restaurants as single large category that a person is choosing from, rather than
modeling preferences across multiple independent categories as we do here.
A second important difference is that the TTFM model predicted choice of restaurant
conditional on the choice to go out to eat, whereas in this paper, we wish to
predict the unconditional purchase probabilities for each product.%
\footnote{However these predictions are still conditioned on the shopper's
  decision to visit the store, which we treat as exogenous.}
That is, we predict both whether or not the shopper will make a purchase from
each product category and if so, which product she will choose.
This is critical for making pricing decisions, since changes to prices may
affect not only which product a customer chooses within a category, but also
whether or not the customer buys anything at all.

Suppose we treated buying nothing from a category, the ``outside good'', as one
more option for the customer to choose from.%
\footnote{i.e.\ we add an additional $U_{i0t} = \epsilon_{i0t}$ to each category,
  representing the decision to buy nothing from the category. Now the set of
  options for the consumer are mutually exclusive, and collectively exhaustive.
  On each shopping trip, for every category, a shopper either chooses something
  from the category or they choose the outside good.}
This, however, makes the assumption of IIA problematic, because the purchase
rates for most product categories in a grocery store are less than 5\%.
If consumer purchases nothing from a category on 95\% of trips, then assuming
IIA implies that if a product is out-of-stock, or its price is increased, this change
will almost never cause a consumer to purchase a different product from the same category.
Instead, most customers will switch to buying nothing from the category, since
that is the most common choice for most customers on most trips.
Under this assumption, because shampoo is purchased relatively infrequently, a price increase for
one brand of shampoo will mostly cause consumers to buy no shampoo at all, with very few consumers
substituting to different choice of shampoo.

To address this concern, we introduce a structure similar to a nested logistic regression.
This gives the model flexibility to fit the degree to which consumers substitute between different
products within a product category rather than deciding to purchase nothing instead.
In this particular application, we use a simple nesting structure with all of
the products in a category in one nest and then a second nest with the outside
good for the category.
However, a similar approach could be used to allow for other more complex nesting structures, since it can be
implemented via repeated runs of the same code used in the TTFM model.
First, a model is trained to predict each customer's choice of product conditional on
the purchase of something from the product's category. The results of this model are aggregated to
calculate a user-category specific ``inclusive value'' term, which is used as an
input into a second run that predicts from which categories each user will make
purchases.
With a more complex nesting structure, a similar process could be repeated in a
bottoms up fashion, with a separate run of the model for each nest, the outputs
of which can be aggregated up to use as inputs for the nests above it.
While this multistage process is statistically inefficient relative to
simultaneous estimation of the full model, it does substantially reduce the
complexity and custom coding required for the computationally intensive portion
of the estimation.%
\footnote{\citet{Train2009} section 4.2.4 provides a nice overview of the
  sequential estimation approach in the context of the traditional nested logit model.}

\subsubsection{Product Choice}
\label{sec:UpcChoice}

To estimate the Nested Factorization model, we first train a model to predict the
consumer's choice of which products they would purchase conditional on the
decision to purchase one item from the corresponding product category.
For example, if the consumer has decided to purchase yogurt---which brand, flavor, and size of yogurt will she
select.
The mapping from utility values to conditional choice probabilities follows the standard multinomial
logit form that arises from assuming a extreme value type 1 distribution for $\epsilon_{ijt}$.
Our model of product choice differs from the standard multinomial logit in that it allows for rich
heterogeneity in preferences and price responsiveness across consumers and
across items.

Similar to the HPF model from \citep{Gopalan2013}, the Nested Factorization model incorporates latent item
characteristics ($\beta_{jk}$) as well as latent user preferences for the latent item
characteristics ($\theta_{ik}$).
The Nested Factorization model extends that model by
incorporating consumer and item level covariates, as well as allowing
time varying characteristics such as price or product availability (e.g.\ a
product being out-of-stock).
These extensions allow for predictions at the level of individual shopping
trips and for predictions of the patterns of substitution between similar
products caused by price changes and changes to product availability.
We assume consumers have latent preferences for observable item characteristics ($\sigma_{ik}$), while
observable user characteristics affect user preferences differentially for each product ($\rho_{jk}$).
We allow for heterogeneity in price elasticities across users and items that
depends on latent item characteristics ($\lambda_j$) and latent user
characteristics ($\gamma_i$).
In addition, we allow for certain items to be out-of-stock or unavailable on a particular
shopping trip ($a_{jt}=0$ if the item is out-of-stock, while $a_{jt}=1$ otherwise).

% Underbraces to explain parts
\begin{equation}
  u_{ijt}
        = \underbrace{\theta_{i}^T \beta_{j}}
            _{\text{Latent - Latent} \atop \text{Intercept}}
        + \underbrace{W_{i}^T \rho_{j}}_{\text{User Observables}}
        + \underbrace{\sigma_{i}^T X_{j}}_{\text{Item Observables}}
        - \underbrace{\gamma_{i}^T \lambda_{j} \text{price}_{jt}}
            _{\text{Latent - Latent} \atop \text{Price Sensitivity}}
\end{equation}

\begin{equation}
  U_{ijt} = u_{ijt} + \epsilon_{ijt}
\end{equation}

\begin{equation} \label{eq:UpcProbs}
  P(y_{ijt} = 1 \mid \sum_{j=1}^{J_c} y_{ijt} = 1)
  = \frac{a_{jt} \exp(u_{ijt})}{\sum_{k=1}^{J_c} a_{jt} \exp(u_{ikt})}
\end{equation}

\subsubsection{Category Choice}
\label{sec:CategoryChoice}

We model the consumer's choice of whether or not to purchase something from each category of goods
as a series of independent binary choices.
The choice to purchase from a category is assumed to depend on the utility values of items in the category, through
their ``inclusive value'' $IV_{ict}$, which is
the expectation (over the realizations of the $\epsilon_{ijt}$) of the maximum
of the utilities of all the available products in the category.%
\footnote{If the $\epsilon$ are a standard Extreme Value Type 1 distribution, then there will
be an extra term $\gamma = \text{mean}(\epsilon) \approx 0.577$ added which in
practice does not matter since it can be absorbed into the constant term.
Alternatively we can define $\epsilon = EV1(-\gamma,1)$ to get rid of the extra
term without affecting any of the choice probabilities.}

\begin{align}
  IV_{ict}
  &\equiv E\left[ \max_{j=1 \ldots J_c} U_{ijt} \right]
  = \log \sum_{j=1}^{J_c} a_{jt} \exp u_{ijt}
    \label{eq:IV}
  \\
  u_{ict}
  &= \underbrace{\vartheta_{i}^T \beta_{c}}
  _{\text{Latent - Latent} \atop \text{Intercept}}
  + \underbrace{W_{i}^T \rho_{c}}_{\text{User Obs.}}
  + \underbrace{\psi_{i}^T X_{c}}_{\text{Category Obs.}}
  - \underbrace{\phi_{i}^T \lambda_{c} IV_{ict}}
  _{\text{Latent - Latent} \atop \text{Inclusive Value}}
    \nonumber
  \\
  &\hphantom{{}=} + \underbrace{\mu_{c} \delta_t}_{\text{Week Trends}}
  + \underbrace{w_{ct}}_{\text{Day of Week}}
    \label{eq:CategoryUtility}
  \\
  U_{ict} &= u_{ijt} + \epsilon_{ict}
    \nonumber
  \\
  P(y_{ict} = 1) &= \frac{\exp(u_{ict})}{1 + \exp(u_{ict})}
  \label{eq:CategoryProbs}
\end{align}

The first three terms of equation \ref{eq:CategoryUtility} capture the user's general
propensity to purchase from this category, which is not affected by the
utilities of any of the products in the category.
The fourth term captures the impact of the inclusive value on the user's choice.
The fifth and sixth terms control for time trends that affect the popularity of
product categories. This helps control for product categories that are systematically more
popular at certain times of year or different days of the week.
The two latent factorization terms allow for rich flexibility in capturing the
correlations in preferences across users.

One interpretation of this model, is that the customer decision of whether or
not to purchase from a category depends in part on her expectation of the
utility she will get from choosing the product that maximizes her utility.
In this interpretation, the $\phi_{i}^T \lambda_c$ coefficients modulate how
much the consumer's choice to buy from the category is affected by her
expectation of utility she would get from picking an item in the category.
If $\phi_{i}^T \lambda_c = 1$, then model simplifies to a non-nested logit\textemdash
IIA holds across all nests.
If $\phi_{i}^T \lambda_c = 0$, then model implies no substitution between
items in different nests\textemdash no price change within the product nest would
ever change your decision of whether or not to buy from the category.

An alternative interpretation of the model frames the nesting structure as
capturing the correlation in the error terms $\epsilon_{ijt}$ of items that are
in the same nest.
In the context of the two group nested structure used in this paper, this
correlation might reflect the consumer's current need for products from this
category, which depends on how much of the product he has at home and what he is
planning on cooking for the week.

\subsubsection{Estimation with Variational Bayes and Stochastic Gradient Descent}
\label{sec:VB}
To estimate the Nested Factorization model, we build on the approach described in
\citet{Ruiz2020} and \citet{Athey2018} to fit a hierarchical Bayesian model to the structure
described in Section~\ref{sec:NF}.
We fit this model as a two step process using the same variational inference code for each step.

In the first step, we estimate the posterior distribution for the latent parameters
$\ell_p = \{\theta_i,\beta_j,\rho_j,\sigma_i,\gamma_i,\lambda_j\}$
that govern each customers conditional purchase probabilities for the products within each
product category.
Conditional on purchasing something from a product category, the model predicts
the choice probabilities for each product within the category.
In each shopping trip, only the categories that a user makes a purchase from are
included in the likelihood.
We denote the product purchase outcomes $\vect{y} = \{y_{ijt}\}$ and the observed data
$\vect{X}$ which includes user characteristics, item characteristics, prices, availability,
and user shopping dates.
We would like to learn the posterior distribution of
the latent parameters $\ell$ across all of the customers, products,
and time periods.
\[
  p(\ell \mid \vect{y}, \vect{x})
  =
  \frac{p(\ell) \prod_t p(y_{ijt} \mid \ell, x_{ijt})}{p(\vect{y} \mid \vect{x})}
\]
As is common in many Bayesian models, the exact posterior distribution over the latent variables
is computationally intractable.
We instead approximate the posterior distribution using variational inference, which is typically
faster on large scale Bayesian problems than classical methods such as
Markov Monte Carlo sampling \citep{Blei2017}.
In variational inference, we select a flexible parameterized family of distributions
$q\left( \ell ; \nu \right)$
over the latent variables of the model $\ell$ and then find the
value of $\nu$ that makes $q\left( \ell ; \nu \right)$ ``close'' to the exact posterior
in terms of Kullback-Leibler divergence.\footnote{KL divergence is similar to
  a distance function in that it is non-negative and $KL(P \mid\mid Q)=0$ iff
  $P=Q$ almost everywhere. It is not a true distance function, however,
  because it is not symmetric $KL(Q \mid\mid P) \neq KL(P \mid\mid Q)$ and does
  not satisfy the triangle inequality.}
In our application, we use a common approach of selecting a mean-field family
for the variational distribution, in which the latent variables are mutually
independent and each is distributed according to a normal distribution with its
own variational parameters for mean and variance.
Minimizing the KL divergence over that possible values of the variational
parameters $\nu$ is equivalent to maximizing what is called the evidence lower
bound (ELBO):
\begin{equation} \label{eq:ELBO}
  \mathscr{L}(\nu) = E_{q(\ell;\nu)}
  \left[
    \log p(\vect{y} \mid \vect{x}, \ell)
    - \log q(\ell; \nu)
  \right]
\end{equation}

Although the expectations that form the ELBO are intractable, we can still seek
to maximize it by noticing that the gradient of the ELBO
$\nabla_\nu \mathscr{L}(\nu)$
can through clever rearrangement be rewritten as the expectation of a tractable formula.%
\footnote{See Appendix \ref{sec:AppendixModel} for more details and
  \citet{Blei2017,Ruiz2020,Athey2018} for additional exposition.}
This means we can use Monte Carlo estimation of the gradient in order to produce
noisy-but-unbiased estimates of the gradient.
We can then use stochastic gradient descent to find the variational parameters
that maximize the ELBO by repeatedly taking small steps in the direction of
these gradient estimates.

The result of this process is an approximation to the posterior
distribution of the latent variables.
We can then use this posterior to obtain a distribution over $u_{ijt}$ for every user, product,
and shopping trip (including for products in categories a user did not make any
purchases from).
We use the means of these estimates to calculate the inclusive value term for
each user and category as in equation~\ref{eq:IV}.\footnote{A downside of this
  two-step approach is that we are not able to take into account the full
  estimated distributions of the latent variables from the product-level model when
  estimating the second category-level model}

We can then repeat the same variational inference process at the category level.
In this stage, each user makes the choice for each category of whether or not to
buy something from the category.
In place of the price term from the first stage, we instead use the predicted
inclusive value term \(IV_{ict} = \log \sum_j \exp u_{ijt}\).
For this stage the latent parameters are
$\ell_c = \{\vartheta_i, \beta_c, \rho_c, \psi_i, \phi_i, \lambda_c, \mu_c, \delta_t, w_{ct}\}$

In order to reuse the variational inference code from the product level
model, we fit the category choice model as if there were two products for each
category: the ``inside good'' with $IV$ calculated based on the products in the
category, and ``outside good'' with $IV = 0$. We then can transform the
estimated parameters by subtracting the outside good's parameters from both,
to be equivalent with a model with the utility of the outside good equal to 0.

In order to make unconditional probabilities of purchase probabilities for each
user, product, and trip, we combine the predictions from the two models.
\begin{equation*}
  P(\text{choose j})
  =
  P(\text{buy from category})
  \cdot
  P(\text{choose j} \mid \text{buy from category})
\end{equation*}

\section{Supermarket Application}

\subsection{Data}
We apply the Nested Factorization model to scanner panel data from one store in a large national
grocery store chain, using a data set originally assembled by \citet{Che2012}.
The data is available to researchers at Stanford and Berkeley by application.
This store is located in an isolated mountain region and has no other large grocery competitors
within a 5 mile radius.
For each transaction that a loyalty-card household makes between May 2005 and March 2007, we observe
the price and quantity of each product purchased.
In addition we incorporate several household demographic variables that the store has compiled from a
variety of sources, including estimated age, gender, income, and household size (there are
additional demographics in our data set but we restricted attention to a subset).
We restrict our analysis to a sample of 2068 households who make between 20 and 300 shopping trips.
These households collectively make 1,551,213 purchases during 333,585 shopping trips.%
% These numbers can be found in poisson-choice/out/Supermarket/Tables/summaryStats.txt
\footnote{We define a shopping trip as a set of all purchases a household makes on a calendar day.}
Of these, 455,445 purchases and 100,504 trips occur on a Tuesday or Wednesday.
We use only the data from Tuesday and Wednesday and exclude weeks with major US
holidays\footnotemark{} in our estimation approach due to concerns about the
potential for price endogeneity as discussed in Section~\ref{sec:unconf}.
\footnotetext{We exclude data from the week prior to Halloween, Thanksgiving, Christmas, 4th of July, and
  Labor Day.}

The data includes a product hierarchy for each product, with the smallest unit of analysis (the unit
at which prices are set) being the universal product code (UPC).
From examining the data, it is not a priori perfectly clear which level of the hierarchy best
matches our desiderata for a ``category,'' which would be for
the consumers to buy at most one item from each category, while purchasing
decisions are not correlated across separate categories.%
\footnote{At higher levels of aggregation, it was much more common to see multiple
  purchases in the same grouping on a single trip. At lower levels of
  aggregation, many categories were split into classes that contained products that
  seemed likely to be substitutes. For example, the category Apples is split
  into classes such as Fuji and Gala apples. Sharp Cheddar is in a separate
  class (but same category) as Mild Cheddar.}
To ensure a good match between the model and the application, we use the ``category'' level of the
UPC hierarchy, and we focus on categories and items that pass certain filters, reducing the number
of product categories from 235 to 123.
Across these categories, a total of 1263 UPCs are included in the sample.
The filters are reviewed in appendix Section~\ref{sec:AppendixData}, but important restrictions
include eliminating highly seasonal categories, as well as categories without sufficient price
variation, or where within-category price changes are highly correlated across products.

\begin{figure}
  \centering
  \includegraphics[width=\fullw]{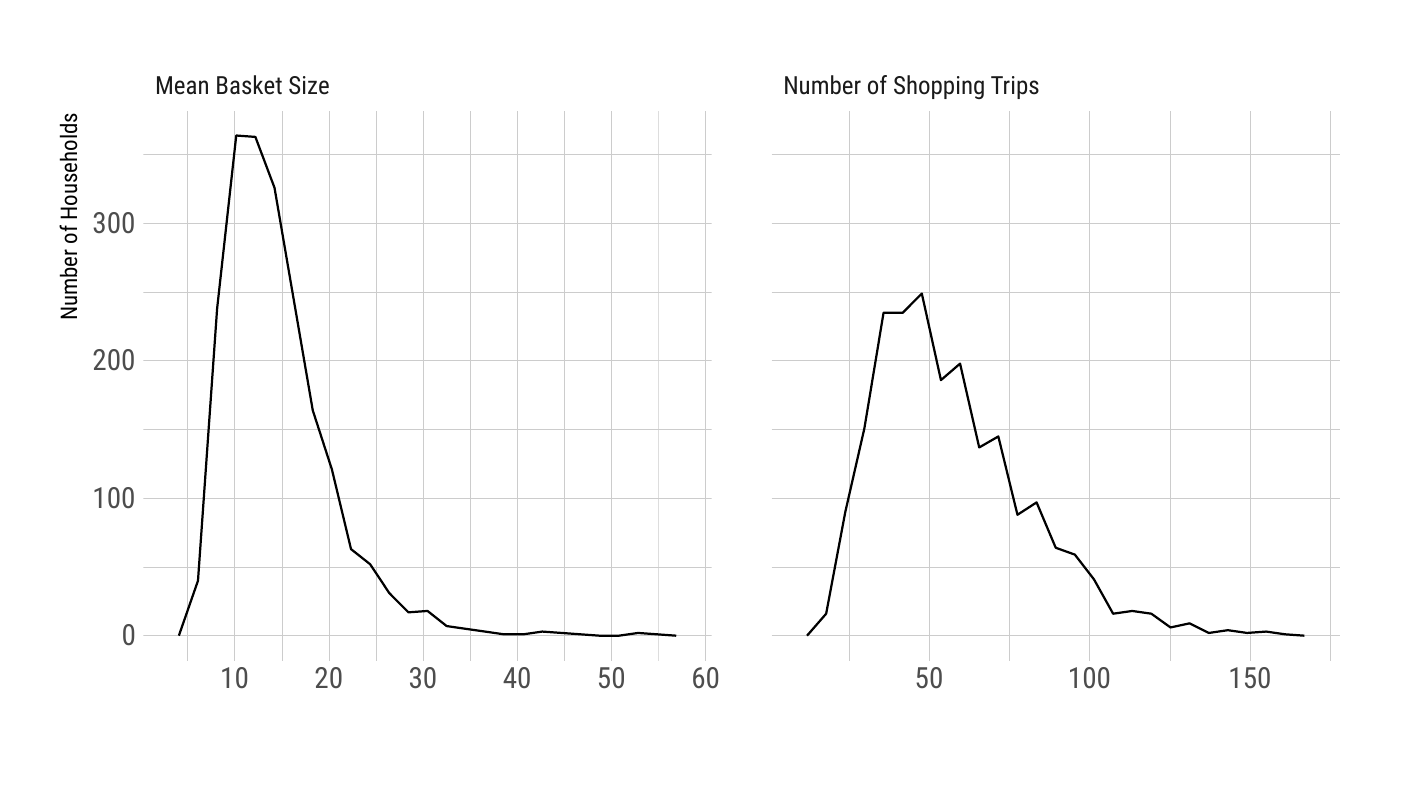}
  \caption{Household Summary Statistics: Tuesday and Wednesday Only}\label{fig:TuesWed}
\end{figure}

Figure~\ref{fig:TuesWed} illustrates summary statistics on household shopping frequency and basket
size in our restricted data set.

\subsection{Models}

\subsubsection{Nested Factorization}
Our primary model is the Nested Factorization model, as outlined in Section~\ref{sec:NF}.
% A core advantage of this model is that it allows us to learn patterns in
% preferences that are correlated across separate product categories. For example
% the type of wine that a shopper often purchases may help us infer which types of
% cheese she is more likely to prefer.
The key hyperparameters of the model are the dimensionality of the latent factorizations of the
user preferences and elasticities.
Allowing for a higher dimensional factorization allows for more flexibility in the shopping patterns
the model is able to fit, at the expense of slower estimation speeds and larger potential for
overfitting the data.
In order to choose the values for these hyperparameters, we follow the standard practice in the
computer science literature of selecting based on performance on a ``validation'' subset of the data
that is distinct from the subset used to train the models (and distinct from the ``test'' subset
that is ``held out'' and not used until the final comparison between models).
We discuss the model selection criteria in more depth in Section~\ref{sec:ModelFit}.
We compare the performance of the Nested Factorization model against several alternative
approaches described in the following sections.

\subsubsection{Multinomial Logit}
The simplest and most commonly used discrete choice model is the multinomial logit
\citep{Train2016}, which has a long history in economics tracing back to \citet{Luce1959} and
\citet{McFadden1974}.

We focus on a baseline specification that controls for household demographics (gender, age, marital
status, and income\footnotemark{}) and include the weekly mean category purchase rates as pseudo-fixed
effects for each calendar week, which helps control for seasonal trends that shift the demand for
the product category and which may be correlated with the product prices.
\footnotetext{We divide age into buckets \{Under 45, 45--55, Over 55\}. We split income at \$100k,
  which is roughly the median for this store.}
We have also tried alternative specifications that add behavioral controls based on splitting
the population into 20\% buckets based on total spending in the store and a
model without demographic controls, all of which had similar predictive performance.
Because of its poor predictive performance across all of the measures we focus
on in this paper, we have omitted the multinomial logit results from some of the
results charts and tables, when the additional entries detract from clarity.

Homogeneous with Demographic Controls:\footnotemark{}
\begin{align*}
  U_{ijt} &= \alpha_{j} + \eta \log p_{jt} + \beta_j D_i + week_t + weekday_t + \epsilon_{ijt}\\
\end{align*}
\footnotetext{In this model and all of the subsequent variations, ``outside
  good'' (the choice to not buy anything in the category, is assumed to take the
  form $U_{i0t} = \epsilon_{ijt}$.}

\subsubsection{Mixed Logit}
The mixed, or random coefficients, logit is one approach for increasing the flexibility of the
multinomial logit.
By allowing the coefficients of the model to vary across the population, the mixed logit allows for
correlation in unobserved factors over time and for more flexible patterns of substitution between
products.
\citet{McFadden2000} show that any choice probabilities derived from random utility maximization can
be approximated arbitrarily well by a appropriately chosen mixed logit model.
As \citet{Steenburgh2013} point out, the mixed logit still constrains demand at the individual
level to satisfy IIA, so it ``improves upon, but does not completely solve the problems of the
homogeneous logit model.''

Mixed with Demographic Controls and Random Price:
\begin{align*}
  U_{ijt} &= \alpha_{j} + \eta_{i} \log p_{jt} + \beta_j D_i + week_t + weekday_t + \epsilon_{ijt}
\end{align*}

Mixed with Random Intercept and Price:
\begin{align*}
  U_{ijt} &= \alpha_{ij} + \eta_{i} \log p_{jt} + week_t + weekday_t + \epsilon_{ijt}
\end{align*}

With sufficiently flexible distributional assumptions for the random
coefficients $\alpha_{ij}$, a mixed logit model could have a similar degree of
flexibility as the latent factorization based approach that we use in the Nested
Factorization model.%
\footnote{For a closer match to the Nested Factorization model, one could
use a random coefficients nested logit model with product-specific random coefficients on
the price in addition to the product intercepts.}
However, scaling such an approach can be computationally challenging, especially
when the number of products is large.
We estimate mixed logit models using the Stata implementation from \citet{Hole2007},
which allows each of the product intercepts $\alpha_{ij}$ to be distributed
across households as independent normal distributions.

\subsubsection{Nested Logit}

Another method for relaxing the homogeneous logit model to allow for more flexible patterns of
substitution is the nested logit model.
In the nested logit model, the decisions a user faces are partitioned into ``nests.''
One interpretation of the nested logit structure is that a user first chooses
which nest to purchase from, and then which product to choose from within the
nest.
An alternative interpretation frames the nesting structure as capturing common
shocks that lead to correlation among the choice probabilities of products in
the same nest.
Within each nest, the choices satisfy the IIA substitution pattern, but the substitution between
products in different nests is able to vary more flexibly.
We choose the same simple nesting structure as was used in the NF model.
We put the outside good in its own nest, and the remaining products in each category in a single
shared nest.
An additional term called the nesting coefficient or inclusive value term controls
how the decision of which nest to choose depends on the utilities of the items within the nest.
This nesting coefficient is analogous to the $\gamma_{i}^T \lambda_{c}$ term in the Nested
Factorization (equation~\ref{eq:CategoryUtility}) with the restriction that the value be homogeneous
across households.
Within the product nest, choice follows the same functional form as used in the homogeneous logit
model with demographic controls.
In essence, the Nested Factorization can be thought of as an extension to a
nested logit that allows for rich heterogeneity in consumer preferences and price sensitivities
by means of a latent factorization approach.

Nested with Demographics:
\begin{align*}
  U_{ijt} &= \alpha_{j} + \eta \log p_{jt} + \beta_j D_i + \epsilon_{ijt} \\
\end{align*}

\subsubsection{Discrete Choice Models with HPF Controls}
\label{sec:HpfFe}

One disadvantage of the Nested Factorization functional form relative to the HPF form used in
\citet{Gopalan2013} is that it leads to substantially slower estimation of the approximate posterior.
The choice of functional form and priors in the original HPF form allows for a closed form for the
gradient of the variational Bayes objective function.
With the Nested Factorization model, we have to perform stochastic gradient
descent using a noisy estimate of the gradient.
In practice this leads the model to require substantially more time and iterations before
convergence.

This motivates an alternative approach that approximates the full Nested Factorization model with a two
step approach.
Estimate each shopper's preferences over items using the HPF model.\footnotemark{}
\footnotetext{We extent the model proposed by \citet{Gopalan2013} to allow for each
  customer to face multiple independent choice occasions, one for each trip they
  make to the store. This extension also allows for time varying characteristics
  such as changes to prices and product availability. However for the
  application presented here, we do not include prices within the HPF model,
  since these outputs are used to plug into models that separately control for prices.}
Then, take the estimated utility values for each shopper and item and plug
these values in as covariates into standard discrete choice models.
This ``HPF controls'' approach can be thought of as an approximation to the generally infeasible
approach of having separate fixed effects for each household item pair (i.e.\ $N \times J$ separate
parameters).

Hierarchical Poisson Factorization (HPF):%
\footnote{Our extension of the original HPF model allows for observed user
and item characteristics (including time varying characteristics), however on
this dataset we found little or no improvement for out of sample predictive fit
relative to a purely latent factorization.}
\begin{align*}
  y_{ijt}   &\sim \text{Poisson}(\mu_{ij}) \\
  \mu_{ij} &=
          \underbrace{\theta_{i}^T \beta_{j}}_{\text{Latent - Latent}}
        + \underbrace{W_{i}^T \rho_{j}}_{\text{User Observables}}
        + \underbrace{\sigma_{i}^T X_{j}}_{\text{Item Observables}}
\end{align*}

This model predicts user $i$ will purchase item $j$ at a mean rate of $\mu_{ij}$,
so we can analogize it to a discrete choice model within a category with utility taking the form
$ u_{ijt} = \log(\mu_{ijt}) + \epsilon_{ijt} $, which will generate approximately the same
choice probabilities for each item in the category $c$.%
\footnote{With the appropriate choice of utility for the outside good
  $u_{i0} = \log{\left( 1 - \sum_{k \in J_c} \mu_{ik} \right)}$, so that
  \(
    P(y_{ijt} = 1)
    = \frac{\exp{u_{ijt}}}{\exp{u_{i0t} + \sum_{k \in J_c} \exp{u_{ikt}}}}
    % = \frac{ \mu_{ijt} }{1 - \sum_{k \in J_c} \exp{u_{ikt}} + \sum_{k \in J_c} \exp{u_{ikt}}}}
    \approx \mu_{ijt}
  \)
  This approximation works best when $\sum_{k \in J_c} \mu_{ik} \ll 1$,
  which is generally true in our supermarket application, but may be less
  appropriate in other contexts where purchase probabilities are larger.
}

This two step procedure may also prove helpful in contexts where it is important to incorporate
additional complications that are difficult to directly embed into the full NF Bayesian model.
For example, it may be effective to include these HPF controls into models of dynamic
discrete choice such as those that arise from storable goods \citep{Hendel2006a} or from consumer
learning \citep{Ackerberg2003a} as a simple way to allow richer heterogeneity in
consumer preferences (at the cost of lower statistical efficiency and potentially bias from
estimating as a two step procedure rather than simultaneously).

This approach may also add value by allowing researchers to take advantage of
data that might otherwise go unused.
Often researchers have access to a broader set of data than the sample they
focus on for their primary analysis.
This might be done in order to take advantage of a subset of the data which was
affected by some source of quasi-experimental variation or may be due to other
desirable properties of a particular subset of users or products.
Latent factorization approaches, such as HPF, may be able to extract useful signal from
the broader set of data, which can then be applied to improve the precision or
flexibility of the primary analysis that focuses on a specific subsample of the data.

\section{Identification and Placebo Tests}
\label{sec:unconf}

In our data, almost all price changes occur on Tuesday nights, near midnight, when very few
customers are shopping.
Thus, we can think of the price change as separating Tuesday and Wednesday.
This motivates an empirical specification in which we use only the data from Tuesday and Wednesday
in order to focus narrowly to the days immediately before and after price changes.
We then include controls at the category level for each week and a indicator variable for Wednesday.
The ``identifying assumption'' for learning price elasticities from this specification is that
any differences in a particular consumer's preferences for items between Tuesday and Wednesday are constant across weeks; in other words, weeks may
differ from one another, but the Tuesday to Wednesday trend is constant over time.
We also exclude the data from weeks immediately prior to major US Holidays out of a concern that
this assumption is less likely to hold in these weeks e.g.\ the difference between the shopping
patterns on the Tuesday and Wednesday before Thanksgiving may systematically differ from pattern
that holds during more typical weeks.
We find in our analysis that the Wednesday effect is small, with the average
difference in category purchase probabilities between Tuesday and Wednesday of
0.127\% (a 3.5\% relative change in purchase probability).

In order to assess the validity of our assumptions, we present some supplementary analysis in the
spirit of the literature on treatment effects \citep{Athey2017State}.
In particular, we test for the presence of certain types of price endogeneity by taking the price
coefficients we get from the actual price data and comparing to the price coefficients we would get
from a model fit on a data with the prices shifted forwards or backwards across time.
We do this in two ways, first by shifting the price of a single UPC in each category (and giving
that UPC a separate price coefficient), and second by simultaneously shifting the prices of all
items in the category.
To create the forward shifted price series for a product, we move each week with a price change
forward to the first week that had no price changes in the real data.%
\footnote{Thus in the shifted data all price changes occur on weeks without price changes in the
  real data and all weeks with price changes in the real data have no price changes in the shifted
  data. A naive shift of all prices by exactly 1 week fails to break the correlation between the
  price changes in the shifted and real price data, due to the frequency of week-long temporary
  price changes.}
We repeat this process for each of the 123 product categories and plot the resulting distribution of
price coefficients and p-values for the price coefficients.  The desired result is that the shifted price
series result in an approximately uniform distribution of p-values, since the artificial price changes should
have no effect on consumer purchase behavior.
The following results are based on the basic multinomial logit specification (with the pseudo week
effects, which are calculated as the category level mean purchase rate in the week).%
\footnote{We focus on the basic multinomial logit specification due to it's computational speed and
  relative simplicity. Running similar tests for the other specifications including the Nested
  Factorization is possible in theory, but requires a larger computational cost.}

Out of the 123 categories, 13 fail one of the four placebo tests at the 1\% level.
If we only considered unconfoundedness checks that shift prices backwards, only 4 categories fail
one of the two backward shifting tests.
Backwards shifts would fail in the presence of consumers who are aware of future price changes.
Forward shifts can fail for goods for which stockpiling is possible.
Some of the categories that fail with the forward shifted price are durable/storable (e.g.\ Baking
Mixes, Ketchup, and Bag Frozen Vegetables), but other categories seem more likely be failing for
other reasons (e.g.\ Refrigerated Turkey and Tomatoes).

\begin{figure}
  \centering
  \begin{minipage}{0.45\textwidth}
    \includegraphics[width=\textwidth]{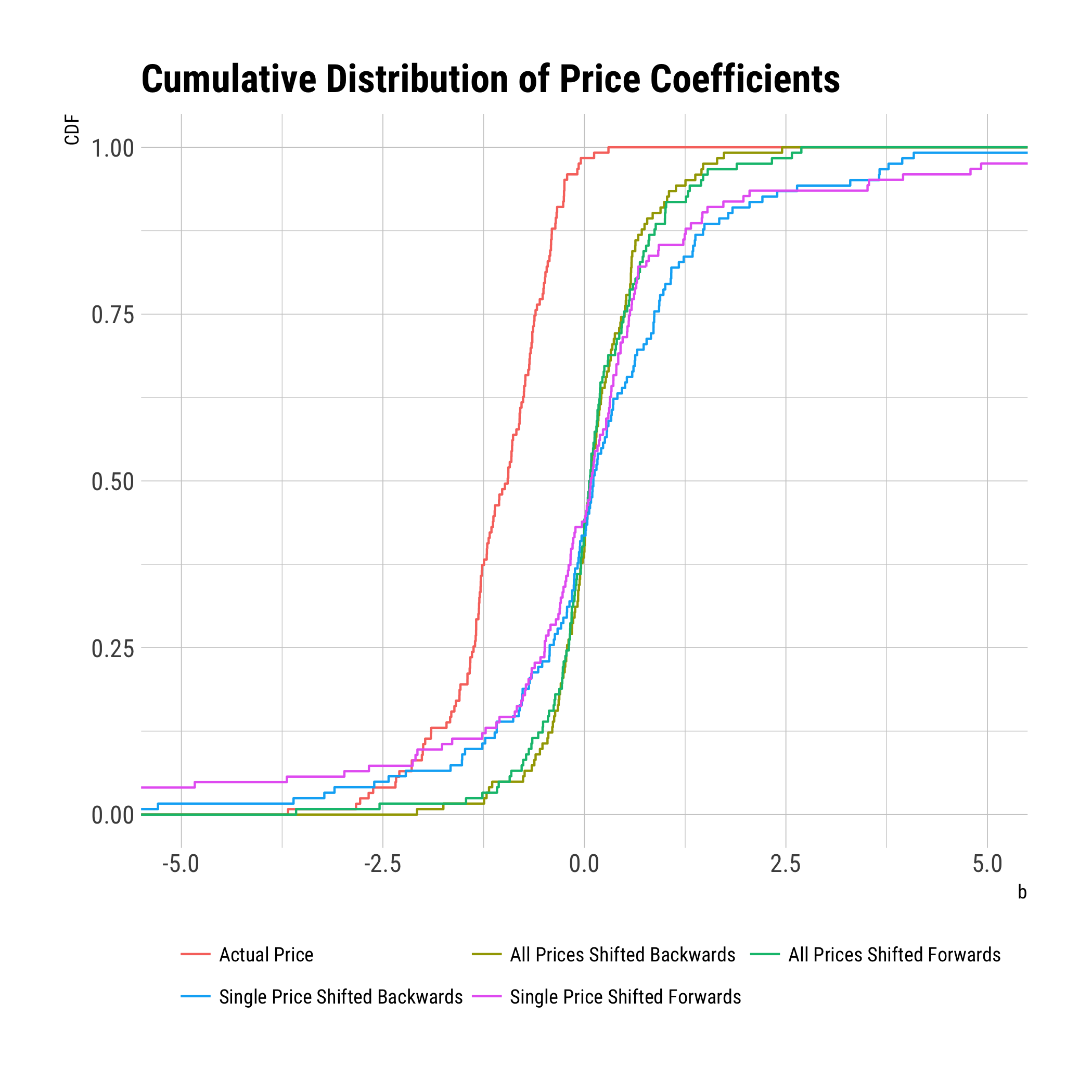}
  \end{minipage}
  \hspace{0.05\textwidth}
  \begin{minipage}{0.45\textwidth}
    \includegraphics[width=\textwidth]{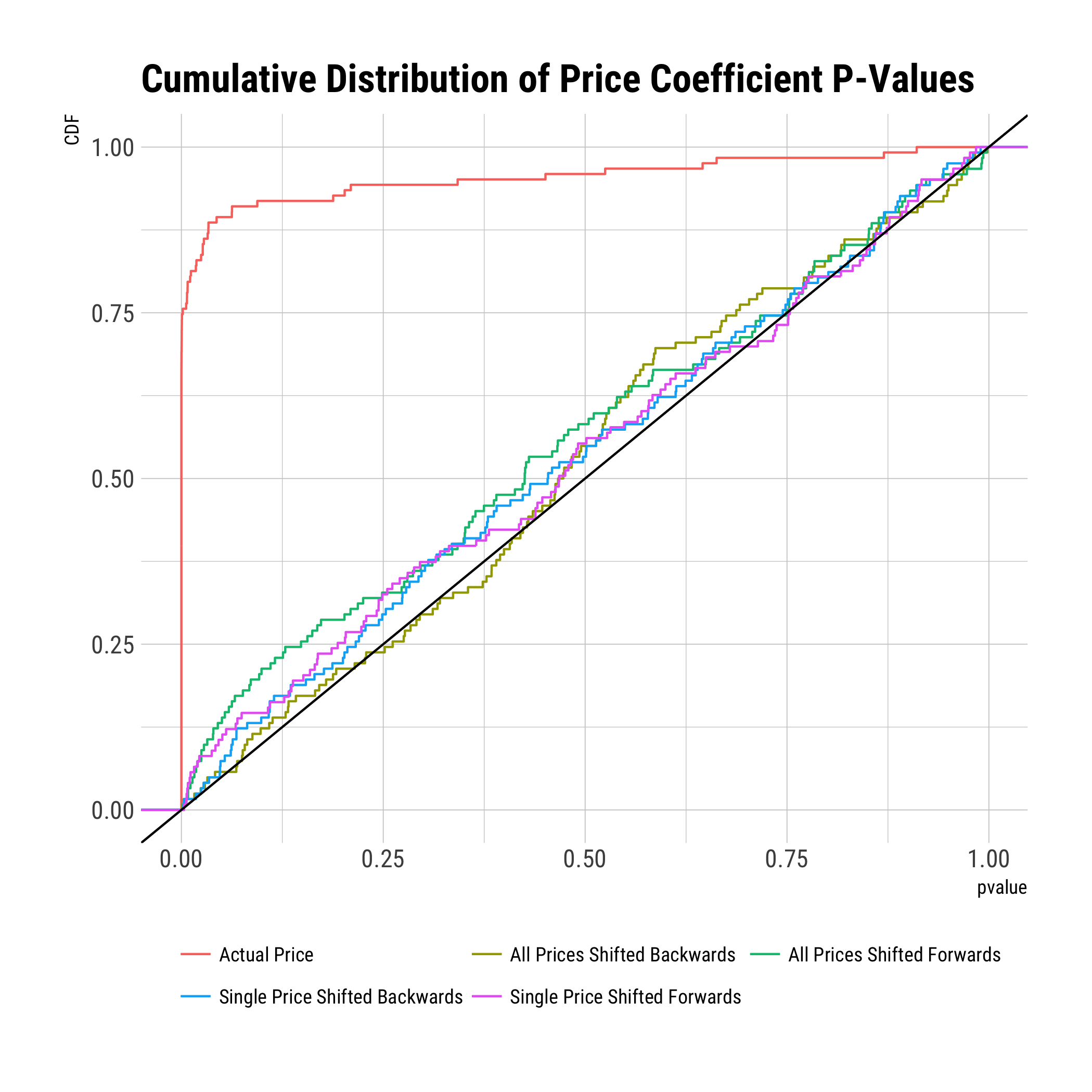}
  \end{minipage}
  \caption{Comparison of price coefficients and p-values with actual price
    series vs. with prices shifted forwards or backwards}\label{fig:Unconfoundedness}
\end{figure}

\section{Assessing Model Performance and Fit}
\label{sec:ModelFit}

In the machine learning literature, it is typical to split data into three non-overlapping parts: a
training set, a validation set, and a test set.
The training data is used to fit the parameters of the model.
To the extent the model has hyperparameters\footnotemark{} that must be set prior to estimation, the
model estimation can be repeated under different values of the hyperparameters.
\footnotetext{For example the regularization coefficient $\lambda$ in a LASSO regression, or in the
  case of Nested Factorization, the number of latent factors of each type to include.}
The validation set is used to select a model (i.e.\ to make a choice of hyperparameters) based on
each model's predictive performance on the validation set.
Finally, the predictive performance on held-out test data is used to evaluate the performance of the
chosen model.
Under the assumption that all observations are drawn from the same data generating process, then the
predictive performance on the test set is an unbiased estimate of the model's ability to make
predictions on new data.
In Section~\ref{sec:PredictiveFit}, we compare Nested Factorization and the set of alternative models
in terms of their predictive fit on the held-out test sample of data.

However, this notion of predictive performance on held-out data does not evaluate the ability of a
model to make causal predictions of what ``would'' happen if we took actions that changed the
distribution of the data.
For example, a model trained to predict the demand for hotel rooms, might correctly identify that
hotels are often full when prices are high and have many empty rooms when prices are low.
This, however, may be due to hotels setting prices in expectation of demand, rather than because
consumers prefer to pay high prices.
Such a model could be highly predictive of hotel demand based on a randomly selected held-out test set (which is drawn
from data generated under the existing data generating process), however such a model would perform
poorly at predicting what prices a hotel ought to charge (since changing prices will change the
data generating process).
It is concerns about such endogeneity of prices%
\footnote{As discussed in \cite{Rossi2014}, with consumer level data, our biggest concern for
  the identification of price effects, is that the store may be setting prices in response to
  variations in expected demand caused by seasonal trends or advertising. For example, there is more
  demand for fresh berries when they are in season or for turkeys immediately before Thanksgiving.
  It is not always clear which direction such price endogeneity will bias our estimates. The
  retailer may decide to take advantage of high demand by raising prices, but in other cases we see
  prices reduced during high demand periods e.g.\ bags of candy going on sale before Halloween.
}
that motivates our identification approach that relies on focusing on data immediately before and
after price changes (Tuesdays and Wednesdays), including weekly time controls (which can absorb any
seasonal/holiday trends), including a indicator variable for Wednesdays (to absorb any consistent
differences in Tuesday vs Wednesday demand), and excluding data from the weeks of major US Holidays.
Since all models include the weekly controls at the category level, they all have the ability to predict
average demand in a week at that level.  However, in weeks with price changes, only a model
that has accurately estimated consumer preferences about price can account for which day within the week
is expected to have more purchases.

To validate the ability to make predictions about counterfactuals, we focus on three types of
changes that can occur during a week for a particular product:
(a) change in the price of the product
(b) a change in the price of a different product in the same category and
(c) another item in the category going into or out of stock.
If the identifying assumptions\footnotemark{} of our models hold, then we can think of each of these
\footnotetext{i.e.\ that controlling for week and day of week effects at the category level is
  sufficient to make potential demand orthogonal to price level and product availability.}
events as a small sources of quasi-experimental variation.
While the outcomes of each of these quasi-experiments is likely to be noisy, we
can improve the precision of our estimates by averaging across a large number of them.

In Section~\ref{sec:counterfactuals}, we compare the log likelihoods of the individual household
level predictions during weeks in which one of these ``counterfactuals'' occurs in order to evaluate
how well each model is able to make predictions that capture the change in predicted demand before
and after the change (relative to the week-level average captured by the weekly time controls).
In addition, we also compare the ability of each model to make predictions about the change in
aggregate demand from Tuesday to Wednesday during weeks in which one of these events occurs.
Our test set holds out data at the household-week level; this allows us to estimate overall consumer
preferences and test our ability to predict household purchases on trips
that were excluded from the training data,
and in particular in weeks where the week-category effect estimated using other consumers' purchases
in that week is insufficient to predict the average probability
of consumers purchasing on a particular day of the week (since prices differ across days).
We use select hyperparameters (tune the model) using only validation set data from item-weeks with price
changes, and we evaluate performance in the test set based on the three changes (a)-(c) outlined above.

In Section~\ref{sec:personalization}, we further evaluate the performance of our model at making
predictions in scenarios of interest for counterfactual inference.  We compare models in terms of their ability to capture
heterogeneity in preferences across the population of households and evaluate the degree to which
the predicted heterogeneity is predictive of actual behavior in the held-out test set.
For example, we compare the predictions made for households who in the training data sample never
purchased a particular UPC or have made no purchases at all from an entire product category.
Among this group of ``never buyers'', we show that the Nested Factorization model is able to
correctly predict which of these households are relatively more or less likely to make a purchase in
the held-out test data.

Section~\ref{sec:elasticities} examines the estimated own-price and cross-price elasticities.
Finally, Section~\ref{sec:target-marketing} looks at the potential for targeted marketing efforts
that are personalized based on the rich heterogeneity estimated by the Nested Factorization model.

\subsection{Predictive Fit}
\label{sec:PredictiveFit}

\begin{table}
  \caption{Comparison of Predictive Fit}\label{tab:PredictiveFit}
  \centering
  \resizebox{\linewidth}{!}{
    \begin{tabular}{lcccc}
      \multicolumn{1}{c}{ } & \multicolumn{2}{c}{Mean Log Likelihood} & \multicolumn{2}{c}{Mean Squared Error} \\
      \cmidrule(l{3pt}r{3pt}){2-3} \cmidrule(l{3pt}r{3pt}){4-5}
      Model & Train & Test & Train & Test\\
      \midrule
      Nested Factorization & -4.2271 & -4.9096 & 0.8981 & 0.9268\\
      Mixed Logit with Random Price and HPF Controls & -4.9233 & -5.3125 & 0.9473 & 0.9660\\
      Nested Logit with HPF Controls & -5.2345 & -5.4230 & 0.9583 & 0.9650\\
      Multinomial Logit with HPF Controls & -5.2307 & -5.4248 & 0.9583 & 0.9651\\
      \addlinespace
      Mixed Logit with Random Price and Demographics & -5.2976 & -5.5690 & 0.9780 & 0.9898\\
      Mixed Logit with Random Price and Random Intercepts & -5.3956 & -5.5827 & 0.9785 & 0.9849\\
      Nested Logit with Demographic Controls & -5.6080 & -5.6779 & 0.9788 & 0.9801\\
      Multinomial Logit with Demographic Controls & -5.6142 & -5.6791 & 0.9791 & 0.9803\\
      \bottomrule
      \bottomrule
      \addlinespace
    \end{tabular}}
\end{table}

In Table~\ref{tab:PredictiveFit}, we compare the predictive fits of each of the models.%
\footnote{Mean Log Likelihood and Mean Squared Error are calculated by dividing
  by the total number of purchases in order to make the values comparable
  between the test and training sets.}
Comparing the overall predictive accuracy across all models, the Nested Factorization model has the
highest likelihood and the lowest sum of squared errors among all models on both the training data
and the held-out test sample.
In addition, each of the models that include HPF controls%
\footnote{i.e.\ user-item specific covariates that are estimated from the HPF model run on all
  categories simultaneously as described in Section~\ref{sec:HpfFe}}
perform better than the models that use control only for demographics.%
\footnote{These trends also hold in additional specifications of the alternative logit models that
  included controls for shopping frequency and previous purchase behavior.}

\subsubsection{Comparison of Predictive Fit by Category}
\label{sec:PredictiveFitByCategory}

We can also compare how each model performs relative to the NF model at the level of individual
categories.
In Table~\ref{tab:CategoryFit}, we calculate the relative rank of each model's performance in the
test set separately for each category.
The Nested Factorization model has the highest log likelihood in 86\% of categories and the lowest
squared error in 96\%.
This demonstrates the effectiveness of learning preferences simultaneously across many product categories.
Even if we were only interested in understanding consumer preferences in one particular category,
e.g.\ yogurt, it can be effective to train a model using the data from other categories as well,
either using the full Nested Factorization model or by using the HPF controls, which also
consistently improve predictive performance on the held-out test data in most categories.

\begin{table}
  \caption{Comparison of Predictive Fit By Category}\label{tab:CategoryFit}
  {\centering
  \resizebox{\linewidth}{!}{
    \begin{tabular}{lcccc}
      \multicolumn{1}{c}{ } & \multicolumn{2}{c}{Mean Rank} & \multicolumn{2}{c}{\% Best Performance} \\
      \cmidrule(l{3pt}r{3pt}){2-3} \cmidrule(l{3pt}r{3pt}){4-5}
      Model & Log L & SE & Log L & SE\\
      \midrule
      Nested Factorization & 1.53 & 1.10 & 86.2\% & 95.9\%\\
      Mixed Logit with Random Price and HPF Controls & 2.40 & 5.33 & 9.8\% & 0.8\%\\
      Nested Logit with HPF Controls & 3.87 & 3.38 & 0.0\% & 0.8\%\\
      Multinomial Logit with HPF Controls & 4.21 & 3.63 & 0.8\% & 0.8\%\\
      \addlinespace
      Mixed Logit with Random Price and Random Intercepts & 4.69 & 5.65 & 0.8\% & 1.6\%\\
      Mixed Logit with Random Price and Demographics & 5.21 & 7.07 & 2.4\% & 0.0\%\\
      Nested Logit with Demographic Controls & 6.96 & 4.93 & 0.0\% & 0.0\%\\
      Multinomial Logit with Demographic Controls & 7.13 & 4.91 & 0.0\% & 0.0\%\\
      \bottomrule
      \bottomrule
      \addlinespace
    \end{tabular}}
  }
  Mean rank is calculated as the average across product categories of the rank
  ordering of models based on predictive fit, with rank 1 corresponding to the
  model with the lowest error.
\end{table}

\subsubsection{Comparison of Fits by Household and UPC}
\label{sec:PredictiveFitByHouseholdAndUpc}

\begin{figure} \centering
  \includegraphics[width=\fullw]{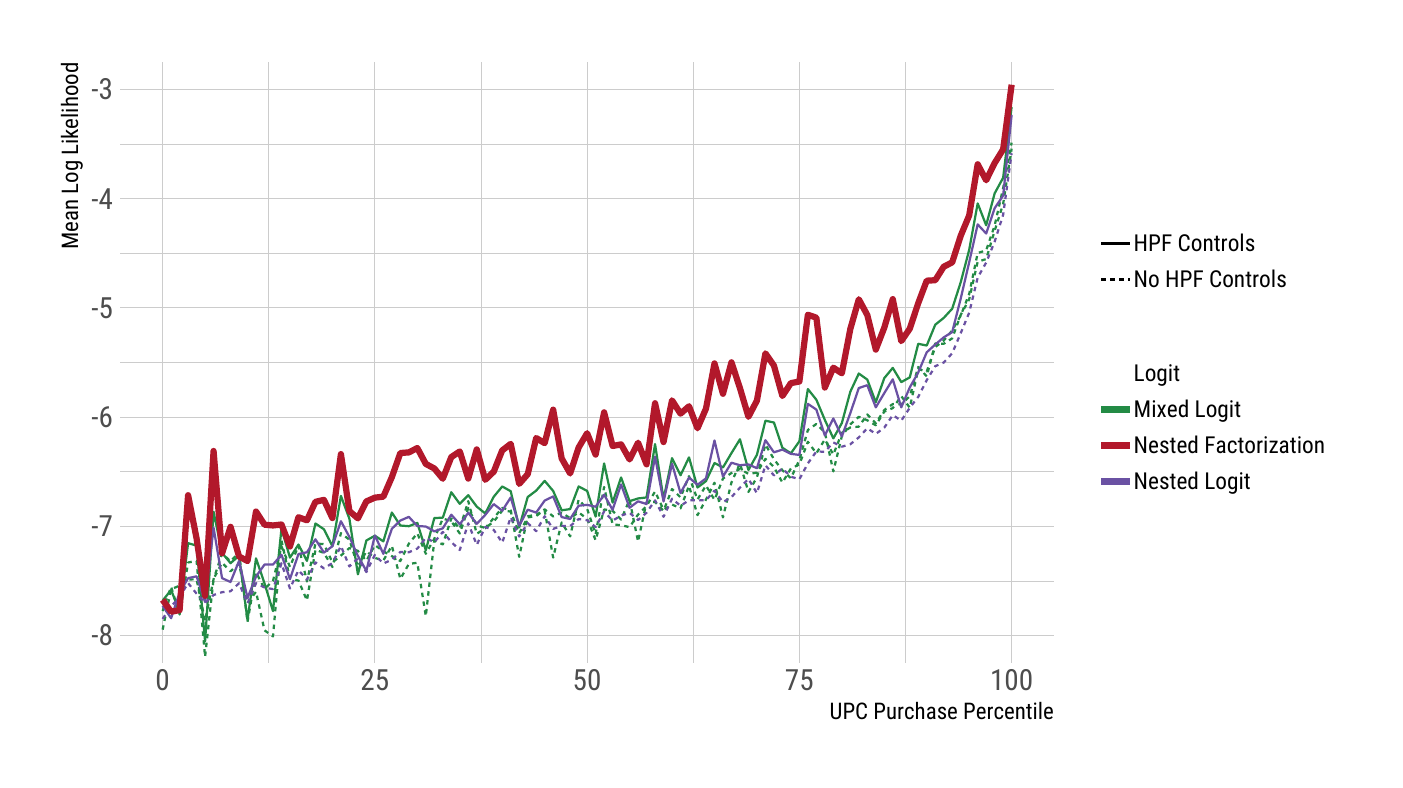}
  \caption{Test Set Predictive Fit by UPC Purchase Frequency}\label{fig:PredictiveFitUpc}
  \includegraphics[width=\fullw]{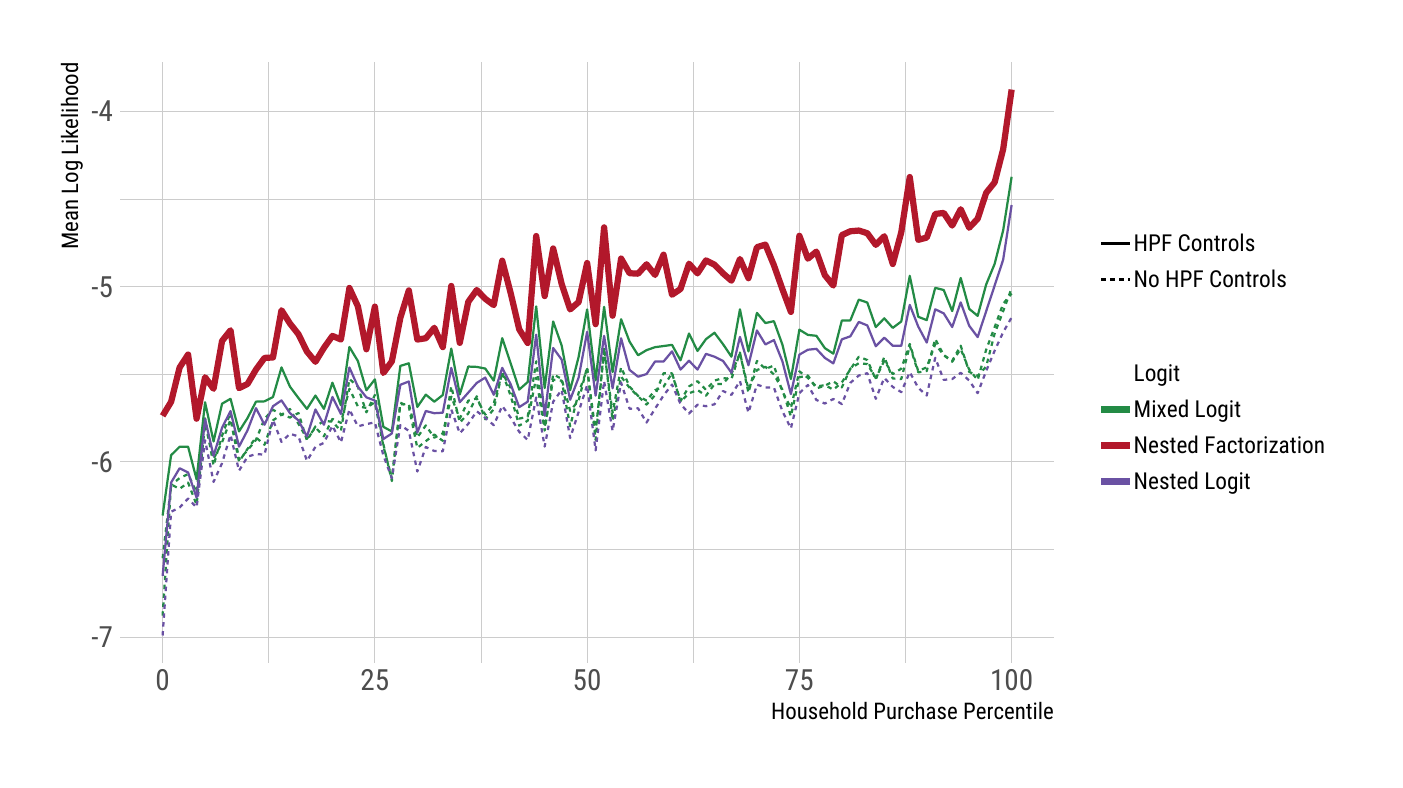}
  \caption{Test Set Predictive Fit by Household Purchase Frequency}\label{fig:PredictiveFitHousehold}
\end{figure}

To help understand where the improvement in predictive performance is coming from,
we can spit the results by the popularity of the products or by the shopping frequency of the households.
In Figure~\ref{fig:PredictiveFitUpc}, we can see that all models are more accurate in their
predictions for the more commonly purchased UPCs (percentile 100) than for the less common items.
Across all percentiles, the Nested Factorization model does consistently better than the alternative
models.
The models that use HPF effects (solid lines) have much smaller, but consistent gains over the
nested and mixed logit models that use demographic or behavioral controls.
Similar trends can be seen when we divide the results based on the number of purchases each
household made in the training data in Figure~\ref{fig:PredictiveFitHousehold}.
This suggests the benefits from a latent-factorization-based approach are relatively uniform across households and products.

\subsection{Price Change and Availability Counterfactuals}
\label{sec:counterfactuals}

In the economics and marketing literatures, models of consumer demand are used to make inferences
about what \emph{would} happen if change were made to a market.
For example, such models have been used to predict what would happen if prices were changed, if
products are added or removed from a market, or if competing firms in the market were to merge.
As discussed in Section~\ref{sec:ModelFit}, evaluating the predictive fit of a model, even when done
on a held-out test sample, does not reliably determine whether a model can be used to make
predictions under counterfactual states of the world such as these.
To evaluate the ability to make predictions under changes to prices or product availability, we
focus on each model's predictions on held-out test set data immediately before and after such
changes occur.
Under the assumption that these changes are exogenous conditional on our week and weekday controls,
we can think of each of these changes as a miniature experiment.
By pooling across many such small noisy experiments, we can increase our precision in detecting
differences in performance.
We focus on three types of changes that can occur between Tuesday and Wednesday for a particular
UPC.
First, we look at weeks in which the focal product's price changes, which we can think of as
evaluating the accuracy of the model's own-price elasticity estimates.
Second, we look at weeks in which some other product in the focal product's category has a price
change, in order to evaluate the predicted cross-price elasticities.
Finally, we look at weeks in which some other product in the focal product's category goes into or
out of stock, which is another measure of the patterns of substitution between products.%
\footnote{In all cases we exclude weeks in which the focal product is out-of-stock on either day.
  For the cross-price and out-of-stock counterfactuals, we exclude weeks in which the focal product
  has a price change. For the price change counterfactuals, we exclude weeks in which the magnitude
  of the price change is less than $\$0.10$}
We asses fit on the test data using three measures.
The first measure is the mean log likelihood of the individual household level predictions for
product weeks that experienced the corresponding counterfactual event.
The second and third measure compare the actual aggregate demand to the predicted aggregate demand
across all households in the test set who shopped during the corresponding weeks.
For products that are purchased at least 2.5 times on average per day, we calculate the likelihood
of the Tuesday to Wednesday change in aggregate demand as approximated by a Skellam distribution.%
\footnote{If the individual purchasing decisions are distributed as independent Bernoulli variables,
  then their sum, the aggregate demand has a Poisson distribution. Then the Tuesday-Wednesday change
  in aggregate demand has a Skellam distribution, which is the difference between two independent
  Poisson distributions}
For less popular products, we calculate the likelihood of observing aggregate demand greater than
zero, which we approximate with a Bernoulli distribution whose mean is the sum of the household
level predictions.

\begin{table}
  \caption{Mean Log Likelihood by Counterfactual Event}\label{tab:Counterfactuals}
  \centering
  \resizebox{\linewidth}{!}{
  \begin{tabular}{lcccc}
  \multicolumn{1}{c}{ } & \multicolumn{2}{c}{Individual} & \multicolumn{2}{c}{Aggregate} \\
  \cmidrule(l{3pt}r{3pt}){2-3} \cmidrule(l{3pt}r{3pt}){4-5}
  Model & Popular & Less Common & Popular & Less Common\\
  \midrule
  \addlinespace[0.3em]
  \multicolumn{5}{l}{\textbf{All Weeks}}\\
  \hspace{1em}Nested Factorization & -0.1070 (0.0004) & -0.0173 (0.0001) & -2.5356 (0.0146) & -1.3072 (0.0044)\\
  \hspace{1em}Mixed Logit with Random Price and HPF Controls & -0.1156 (0.0004) & -0.0188 (0.0001) & -2.5562 (0.0145) & -1.3365 (0.0037)\\
  \hspace{1em}Nested Logit with HPF Controls & -0.1194 (0.0005) & -0.0191 (0.0001) & -2.5568 (0.0154) & -1.3105 (0.0040)\\
  \hspace{1em}Multinomial Logit with HPF Controls & -0.1194 (0.0005) & -0.0191 (0.0001) & -2.5554 (0.0154) & -1.3086 (0.0040)\\
  \hspace{1em}Mixed Logit with Random Price and Random Intercepts & -0.1263 (0.0005) & -0.0194 (0.0001) & -2.5712 (0.0146) & -1.3316 (0.0037)\\
  \hspace{1em}Mixed Logit with Random Price and Demographics & -0.1240 (0.0004) & -0.0195 (0.0001) & -2.5934 (0.0153) & -1.3543 (0.0038)\\
  \hspace{1em}Multinomial Logit with Demographic Controls & -0.1290 (0.0005) & -0.0197 (0.0001) & -2.5722 (0.0157) & -1.3105 (0.0040)\\
  \hspace{1em}Nested Logit with Demographic Controls & -0.1289 (0.0005) & -0.0197 (0.0001) & -2.5740 (0.0157) & -1.3113 (0.0041)\\
  \addlinespace[0.3em]
  \multicolumn{5}{l}{\textbf{Cross Price Weeks}}\\
  \hspace{1em}Nested Factorization & -0.0925 (0.0008) & -0.0149 (0.0001) & -2.4527 (0.0262) & -1.2017 (0.0084)\\
  \hspace{1em}Mixed Logit with Random Price and HPF Controls & -0.1006 (0.0008) & -0.0163 (0.0001) & -2.4846 (0.0258) & -1.2584 (0.0069)\\
  \hspace{1em}Nested Logit with HPF Controls & -0.1041 (0.0009) & -0.0164 (0.0001) & -2.4844 (0.0277) & -1.2177 (0.0076)\\
  \hspace{1em}Multinomial Logit with HPF Controls & -0.1041 (0.0009) & -0.0164 (0.0001) & -2.4836 (0.0276) & -1.2165 (0.0076)\\
  \hspace{1em}Mixed Logit with Random Price and Random Intercepts & -0.1139 (0.0009) & -0.0168 (0.0001) & -2.5002 (0.0260) & -1.2527 (0.0069)\\
  \hspace{1em}Mixed Logit with Random Price and Demographics & -0.1111 (0.0009) & -0.0170 (0.0001) & -2.5132 (0.0280) & -1.2752 (0.0071)\\
  \hspace{1em}Multinomial Logit with Demographic Controls & -0.1162 (0.0010) & -0.0170 (0.0001) & -2.4966 (0.0279) & -1.2182 (0.0077)\\
  \hspace{1em}Nested Logit with Demographic Controls & -0.1161 (0.0010) & -0.0170 (0.0001) & -2.5066 (0.0292) & -1.2194 (0.0078)\\
  \addlinespace[0.3em]
  \multicolumn{5}{l}{\textbf{Own Price Weeks}}\\
  \hspace{1em}Nested Factorization & -0.1374 (0.0009) & -0.0229 (0.0001) & -2.7871 (0.0360) & -1.5544 (0.0104)\\
  \hspace{1em}Mixed Logit with Random Price and HPF Controls & -0.1465 (0.0009) & -0.0243 (0.0001) & -2.8004 (0.0357) & -1.5475 (0.0086)\\
  \hspace{1em}Nested Logit with HPF Controls & -0.1493 (0.0009) & -0.0249 (0.0001) & -2.7986 (0.0371) & -1.5356 (0.0092)\\
  \hspace{1em}Multinomial Logit with HPF Controls & -0.1493 (0.0009) & -0.0249 (0.0001) & -2.7949 (0.0373) & -1.5321 (0.0092)\\
  \hspace{1em}Mixed Logit with Random Price and Random Intercepts & -0.1530 (0.0009) & -0.0251 (0.0001) & -2.8003 (0.0348) & -1.5408 (0.0085)\\
  \hspace{1em}Mixed Logit with Random Price and Demographics & -0.1525 (0.0009) & -0.0251 (0.0001) & -2.8544 (0.0374) & -1.5709 (0.0090)\\
  \hspace{1em}Multinomial Logit with Demographic Controls & -0.1557 (0.0010) & -0.0256 (0.0002) & -2.8097 (0.0369) & -1.5344 (0.0092)\\
  \hspace{1em}Nested Logit with Demographic Controls & -0.1555 (0.0010) & -0.0256 (0.0002) & -2.8186 (0.0372) & -1.5377 (0.0093)\\
  \addlinespace[0.3em]
  \multicolumn{5}{l}{\textbf{Out of Stock Weeks}}\\
  \hspace{1em}Nested Factorization & -0.0924 (0.0040) & -0.0159 (0.0002) & -2.3349 (0.1114) & -1.2746 (0.0129)\\
  \hspace{1em}Mixed Logit with Random Price and HPF Controls & -0.1033 (0.0044) & -0.0173 (0.0002) & -2.3679 (0.1225) & -1.3064 (0.0111)\\
  \hspace{1em}Nested Logit with HPF Controls & -0.1068 (0.0046) & -0.0176 (0.0002) & -2.3427 (0.1243) & -1.2817 (0.0122)\\
  \hspace{1em}Multinomial Logit with HPF Controls & -0.1068 (0.0046) & -0.0176 (0.0002) & -2.3446 (0.1259) & -1.2774 (0.0121)\\
  \hspace{1em}Mixed Logit with Random Price and Demographics & -0.1091 (0.0045) & -0.0180 (0.0002) & -2.3669 (0.1132) & -1.3249 (0.0118)\\
  \hspace{1em}Mixed Logit with Random Price and Random Intercepts & -0.1125 (0.0046) & -0.0181 (0.0002) & -2.3403 (0.1156) & -1.3068 (0.0113)\\
  \hspace{1em}Multinomial Logit with Demographic Controls & -0.1146 (0.0048) & -0.0183 (0.0002) & -2.3261 (0.1189) & -1.2800 (0.0122)\\
  \hspace{1em}Nested Logit with Demographic Controls & -0.1145 (0.0047) & -0.0183 (0.0002) & -2.2974 (0.0990) & -1.2840 (0.0123)\\
  \bottomrule
  \bottomrule
  \addlinespace
  \end{tabular}}
\end{table}

\subsection{Comparison of Degree of Personalization across Households}
\label{sec:personalization}
To examine the extent to which each of these models is able to flexibly model the differences in
preferences between households, we calculate two measures of the degree of ``personalization'' of
the predicted purchase rates that each model predicts for each household.
First, we compare the coefficient of variation of a models predictions at the UPC level and the
category level.%
\footnote{Coefficient of variation is defined as $\frac{sd}{mean}$}
As a second measure, we regress the predicted purchase rate on the actual purchase rate in the
in the held-out test sample.%
\footnote{The coefficients are from a regression of actual purchase rate on the predicted purchase
  rate (both calculated on the test set) with item/category specific fixed effects to absorb heterogeneity in
  the mean purchase rates across items/categories.}

Table~\ref{tab:Personalization} shows that the Nested Factorization model has the largest variation
in the predictions across households and that this variation is strongly correlated with variation
in the actual purchase rates in the held-out test data.
For each $1\%$ increase in the Nested Factorization model's prediction of a household's purchase for
a UPC, the household's actual purchase rate in the test set increases by $0.9955\%$.

\begin{table}
  \caption{Comparison of Degree of Personalization of Predictions across Models}\label{tab:Personalization}
  \centering
  \resizebox{\linewidth}{!}{
    \begin{tabular}{lcccc}
      \multicolumn{1}{c}{ } & \multicolumn{2}{c}{Coef. of Variation} & \multicolumn{2}{c}{Regression Coef.} \\
      \cmidrule(l{3pt}r{3pt}){2-3} \cmidrule(l{3pt}r{3pt}){4-5}
      Model & UPC & Category & UPC & Category\\
      \midrule
      Nested Factorization & 3.2546 & 1.7756 & 0.9955 & 1.0023\\
      Mixed Logit with Random Price and HPF Controls & 2.0747 & 1.6085 & 0.6861 & 0.7007\\
      Mixed Logit with Random Price and Demographics & 1.3869 & 1.5724 & 0.4718 & 0.5968\\
      Multinomial Logit with HPF Controls & 1.2590 & 0.7276 & 0.8402 & 0.8893\\
      \addlinespace
      Nested Logit with HPF Controls & 1.2368 & 0.7520 & 0.8417 & 0.8725\\
      Mixed Logit with Random Price and Random Intercepts & 1.0834 & 1.0446 & 0.4666 & 0.6959\\
      Nested Logit with Demographic Controls & 0.4465 & 0.2967 & 0.8947 & 0.9314\\
      Multinomial Logit with Demographic Controls & 0.4337 & 0.2756 & 0.9077 & 0.9411\\
      \bottomrule
      \bottomrule
      \addlinespace
    \end{tabular}}
\end{table}

\subsubsection{Predicting Preference for Products a Household has Not Yet Purchased}

\begin{figure} \centering
  \includegraphics[width=\fullw]{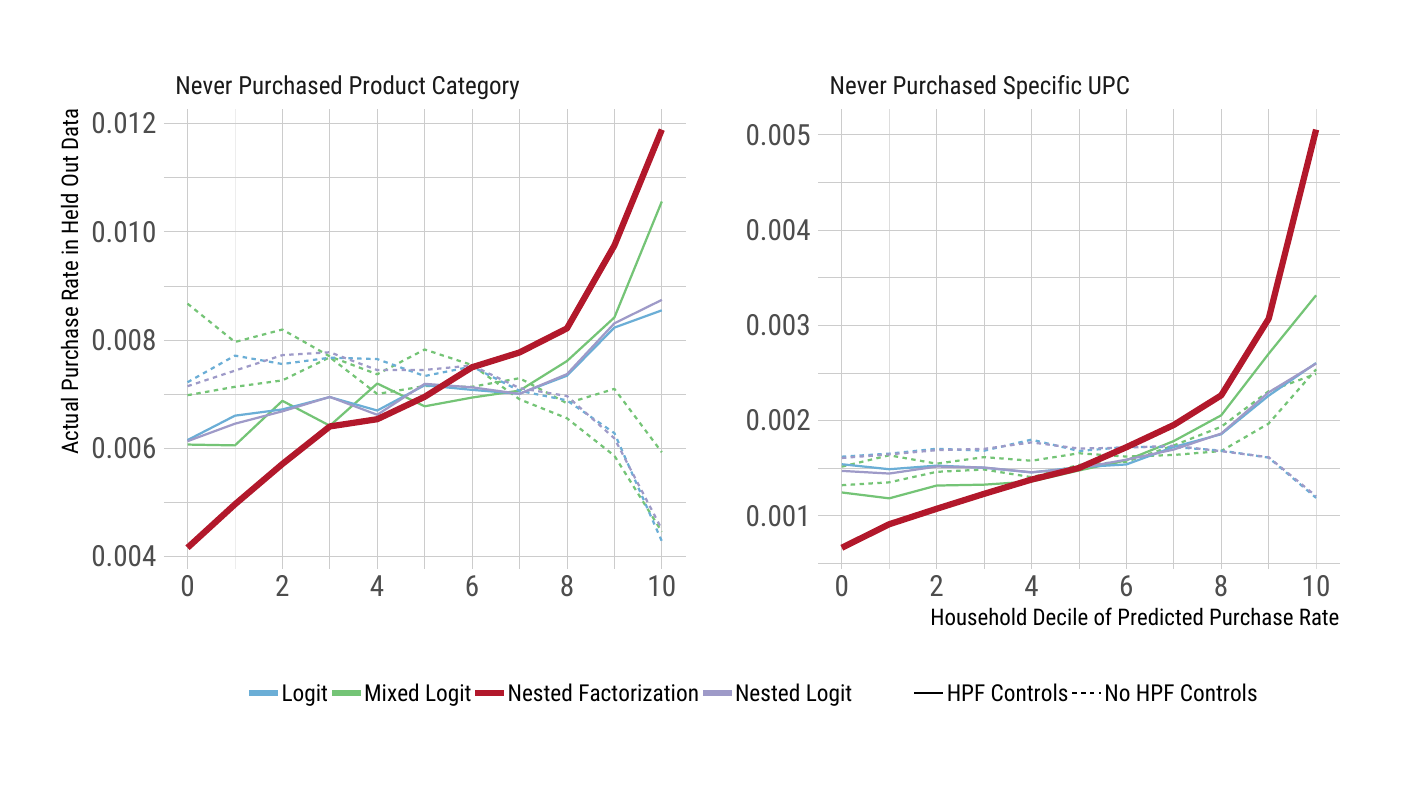}
  \caption{True purchase rate in held out test set for households who never purchased the UPC/Category in the training sample.}\label{fig:NeverBuyerUpcAndCategory}
\end{figure}

The Nested Factorization model is also able to make predictions about the strength of a households
preferences for a given UPC, even if the household has never purchased that particular item before.
To demonstrate this, we look at the set of all households who have made 0 purchases of a particular
UPC in the training sample.
For each UPC and each predictive model, we can rank these households based on their predicted
purchase rate and group them into deciles.
We carry out a similar analysis of households who made no purchases from an entire product category
during the training sample.
Figure~\ref{fig:NeverBuyerUpcAndCategory} shows that the Nested Factorization model is able to correctly
predict which households are relatively more or less likely to purchase a category or UPC that they
never purchased during the training sample.
The decile of households with the highest predicted likelihood to purchase a product category for
the first time, purchases at roughly 3 times the frequency of the lowest decile.
At the UPC level, this ratio of purchase rates in the held-out test sample is more than 10 fold difference.
The models with HPF controls (solid lines) are able to capture a smaller amount of this variation
The models without either approach for latent factorization (dotted lines) have substantially less
predictive power for these first time buyers.
This may be useful to applied marketing practitioners who may be interested in targeting advertising
or promotions towards new customers who might be interested in a product that they have not yet tried.

\subsection{Estimated Elasticities}
\label{sec:elasticities}

\subsubsection{Cross-Price Elasticities Within and Between Products Subcategories}

A different approach for assessing models is to see how well each does at
learning patterns that we believe to be true about the world.
None of the models were given data from the product hierarchy data about the ``class'' or
``subclass'' each product is categorized under.
These groupings are more fine grained than the ``category'' level that we focused on for modeling
product substitution.
Nevertheless, we show that the Nested Factorization model correctly infers that products that are in
the same class or subclass are more similar to each other, and as a result these models predict
higher levels of cross-price elasticities between products that are in the same class or subclass
than between items that are in different classes/subclasses.

\begin{table}
  \caption{Comparison of Cross-Price Elasticities}\label{tab:Elasticities}
  {\centering
  \resizebox{\linewidth}{!}{
  \begin{tabular}{lccccccccc}
  \multicolumn{1}{c}{ } &  \multicolumn{3}{c}{Class Cross Price} & \multicolumn{3}{c}{Subclass Cross Price} \\
  \cmidrule(l{3pt}r{3pt}){2-4} \cmidrule(l{3pt}r{3pt}){5-7}
  Model                                               & Inside & Outside & \%    & Inside & Outside & \%\\
  \midrule
  Nested Factorization                                & 0.0186 & 0.0080  & 132\% & 0.0196 & 0.0181  & 8.4\%\\
  Nested Logit with Demographic Controls              & 0.0119 & 0.0086  & 37\%  & 0.0125 & 0.0115  & 8.5\%\\
  Mixed Logit with Random Price and Random Intercepts & 0.0062 & 0.0053  & 16\%  & 0.0062 & 0.0062  & -0.7\%\\
  Mixed Logit with Random Price and HPF Controls      & 0.0063 & 0.0054  & 16\%  & 0.0063 & 0.0065  & -3.9\%\\
  Multinomial Logit with HPF Controls                 & 0.0035 & 0.0030  & 16\%  & 0.0036 & 0.0035  & 0.8\%\\
  Multinomial Logit with Demographic Controls         & 0.0034 & 0.0029  & 16\%  & 0.0034 & 0.0034  & 2.1\%\\
  Mixed Logit with Random Price and Demographics      & 0.0067 & 0.0060  & 13\%  & 0.0066 & 0.0069  & -4.6\%\\
  Nested Logit with HPF Controls                      & 0.0148 & 0.0158  & -7\%  & 0.0140 & 0.0157  & -10.7\%\\
  \bottomrule
  \bottomrule
  \addlinespace
  \end{tabular}}
  }
    We compare the mean estimated elasticities
    for products that are/aren't in the same product class or product subclass.
    We would expect the cross-price elasticities inside of a class or subclass
    to be higher than those outside the class, since this implies substitution
    towards products that are more similar.
\end{table}

\subsubsection{Aggregate Demand Curves}
\begin{figure}
  \centering
  \includegraphics[width=\fullw]{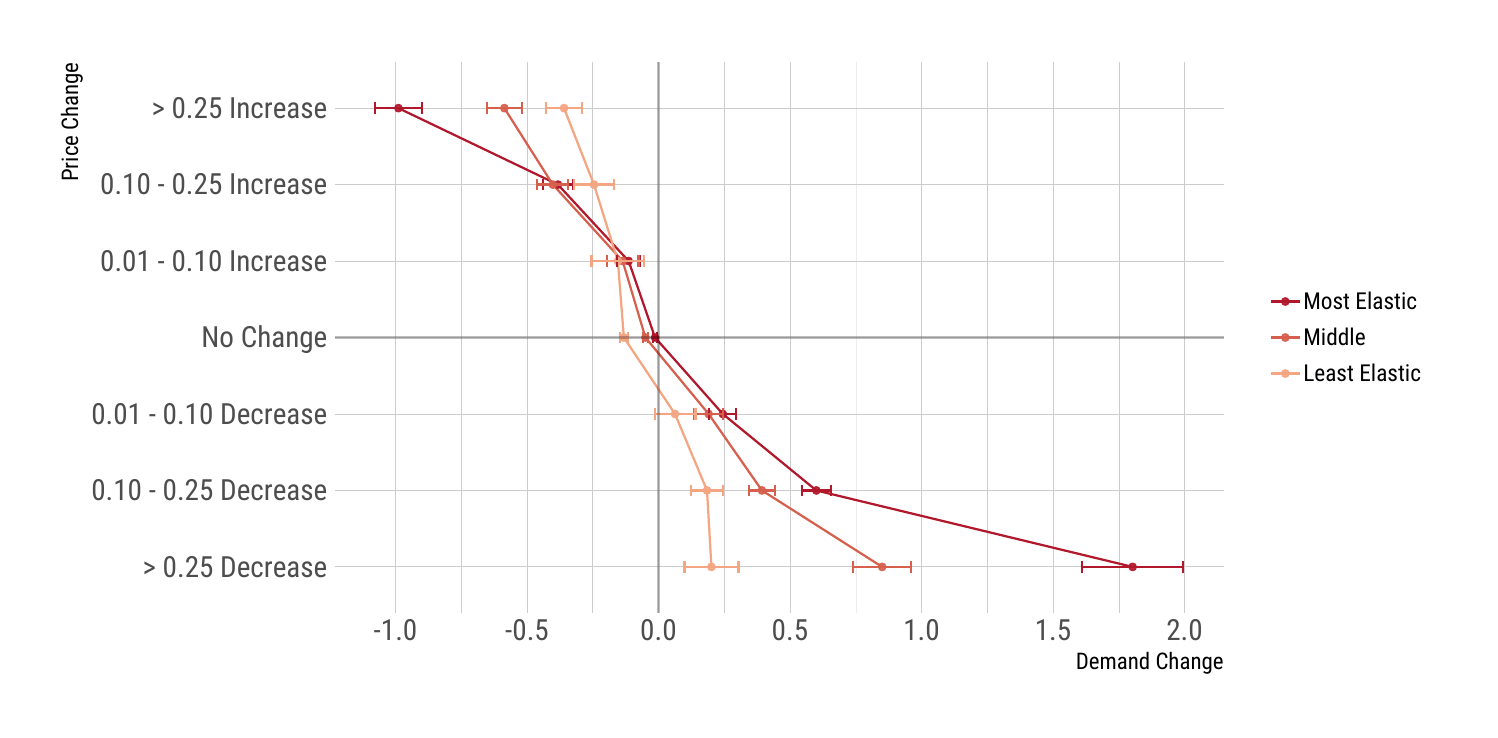}
  \caption{Test Set Aggregate Demand Change by Elasticity Tercile
    }\label{fig:ElasticityDemandChanges}
\end{figure}

In Figure~\ref{fig:ElasticityDemandChanges}, we validate whether the households with higher
predicted elasticities do in fact respond more to price changes in held-out test data.
To do this, for each UPC we split households into terciles based on their predicted elasticity.
We then compare the Tuesday-Wednesday change in aggregate demand in the test set depending on the
size of the price change during the week.
The household's with the higher predicted elasticities do in fact appear to have aggregate demand
that is more responsive to price changes.  This establishes that the heterogeneity we estimate is useful
for counterfactual predictions about heterogeneity in consumer response to price changes.

\subsection{Target Marketing}
\label{sec:target-marketing}

Understanding how preferences vary across customers can serve as an important input into the decision
of how to allocate resources across marketing efforts such as advertising and coupons.
In Table~\ref{tab:CouponTargeting}, we carry out an analysis similar to the classic study
by \citet{Rossi1996} to estimate the potential impact of targeting coupons to customers
based on their purchase histories.
For each product category, we use each model to select the 30\% of households for whom it would be
most profitable for the store to offer a 30\% off coupon for the most popular UPC in the category.%
\footnote{Profits are calculated as price - marginal cost. Marginal costs come from the retailer's
  records, which are available for most products. For items with no marginal cost data, we treat
  the minimum retail price in the data as the marginal cost.}
We then evaluate how profitable this coupon targeting would be using the Nested Factorization model
as ground truth.
For each model, we evaluate three approaches to targeting the coupons.
Under individualized targeting, the store is able to select individual households when choosing whom
to target with the coupons.
Under demographic targeting, the coupons must be allocated in a way that is uniform within
demographic groups.%
\footnote{We define demographic groups in terms of marital status, income level, age, and number of children.}
Under behavioral targeting, the coupons must be allocated based on the number of times a household
has made purchases in the product category.
Under each scenario, we compare the predicted store profits to the profits that would have been
earned if the store had allocated the coupons uniformly at random.

\begin{table}
  \caption{Gains from Targeted Discounts:}\label{tab:CouponTargeting}
  \centering
  \resizebox{\linewidth}{!}{
    \begin{tabular}{lccc}
      \multicolumn{1}{c}{ } & \multicolumn{3}{c}{\% Gains Relative to Uniform} \\
      \cmidrule(l{3pt}r{3pt}){2-4}
      Model & Behavioral & Demographic & Individualized\\
      \midrule
      Nested Factorization (linear) & 2.57\% & 4.55\% & 28.5\%\\
      Mixed Logit with Random Price Effects and HPF Controls (linear) & 1.51\% & 2.38\% & 5.7\%\\
      Multinomial Logit with HPF Controls (linear) & 1.37\% & 1.60\% & 5.7\%\\
      Nested Logit with HPF Controls (linear) & 1.41\% & 1.57\% & 5.4\%\\
      Nested Logit with Demographic Controls (linear) & 0.56\% & 3.03\% & 4.7\%\\
      \addlinespace
      Multinomial Logit with Demographic Controls (linear) & 0.77\% & 2.17\% & 3.4\%\\
      Mixed Logit with Random Price Effects and Demographics (linear) & 0.75\% & 2.53\% & 3.1\%\\
      Mixed Logit with Random Price and Random Intercepts (linear) & 0.86\% & 1.31\% & 2.5\%\\
      \bottomrule
      \bottomrule
      \addlinespace
    \end{tabular}}
\end{table}

\begin{figure}
  \centering
  \includegraphics[width=\midw]{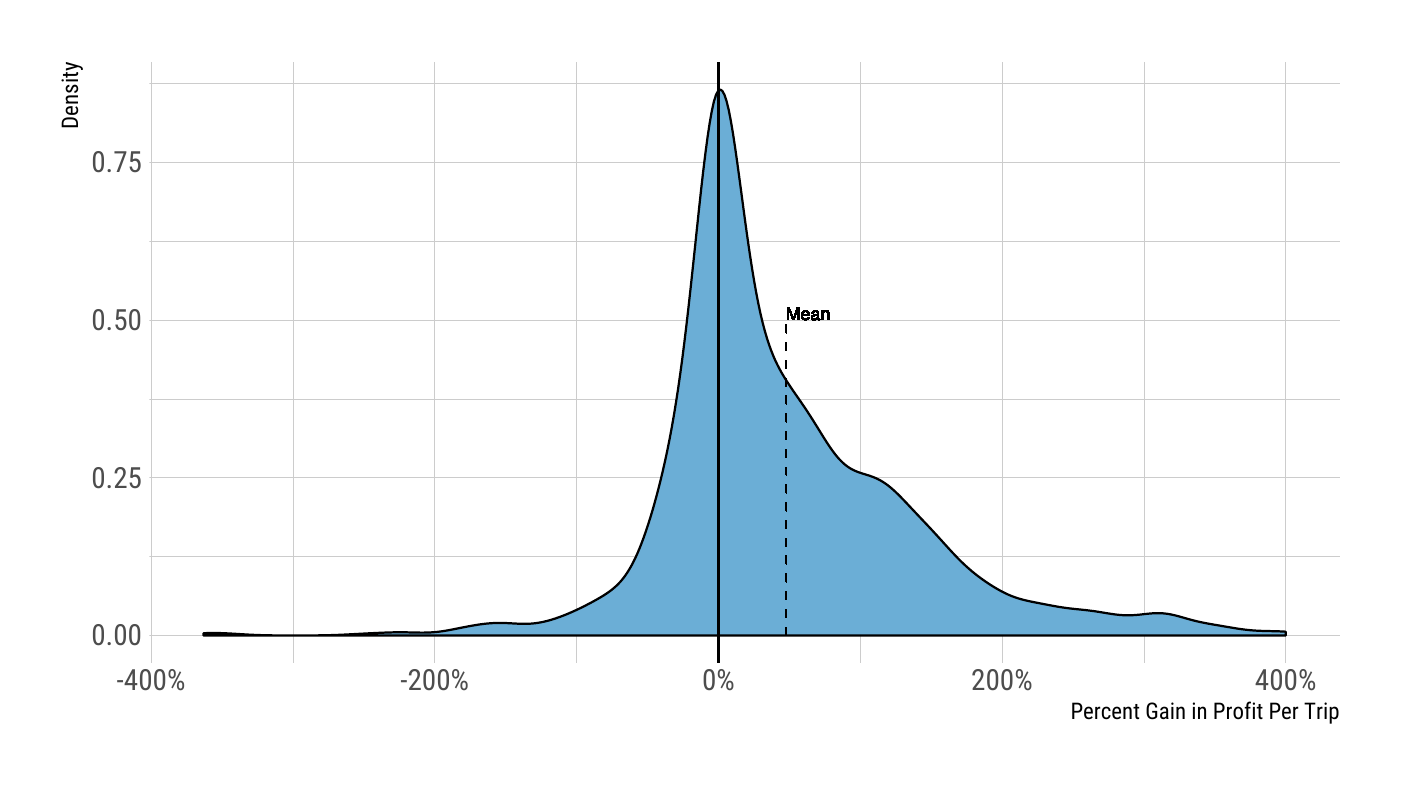}
  \caption{Distribution of Gain in Profit from Targeted Price\\
    Percent gain calculated as
    $\frac{\text{Difference in Average Profit}}{\text{Average Profit}}$
    for each item and price group, where the difference in average profit is the mean profit per
    shopping trip under the preferred price minus the mean profit per shopping trip under the
    alternative price.}\label{fig:PersonalPriceDistribution}
\end{figure}

Unfortunately, without the ability to run an experiment, it is difficult to validate our model's
predictions of the household specific profitability of pricing decisions.
We can however approximate such an experiment by looking at consumers' purchasing behavior under the
various price regimes that happened to have occurred during each consumer's test sample shopping
trips.
\footnote{The validity of this analysis requires that customers are not
  strategically choosing which days to shop in response to prices. In our
  context with prices frequently changing across many categories, we believe
  that this effect is small. However, it is possible that some customers
  might be able to time their shopping trips in response to the prices of a few
  products that they consider particularly important.}
For each UPC, we identify the two most common prices\footnotemark{}, and for each consumer use our
model to predict which of the prices will lead to higher store profits.
\footnotetext{We exclude all prices that are less than the item's marginal cost, since those prices
  would lead to negative profits, and thus would never be chosen as the more profitable price for
  any consumer.}
We can then compare the average profit per shopping trip from the focal UPC under the two chosen
prices.
We aggregate these profits across the two groupings of households%
\footnote{i.e.\ the household's who are predicted to have higher profits under price 1 and the
  households who are predicted to have higher profits under price 2.}
and calculate the increase in average profits per shopping trip from households shopping at their
targeted price relative to the other price.%
\footnote{We restrict ourselves to the two most common prices in order to increase the frequency
  with which we observe shopping trips with the selected prices in the test sample.}
In Figure~\ref{fig:PersonalPriceDistribution}, we can see that on average the store earns
substantially more profit from households when they shop on days with the price that we predicted
would lead to higher profits.
In principal, a similar approach could be used to evaluate a model's predictions
about which users are most likely to switch brands when a new product enters a
category.
Unfortunately in our application, the sample selection criteria we used dropped
most products that entered or exited the market during the sample period.

\section{Conclusion}
This paper proposes the Nested Factorization model for learning consumer preferences from panel
data.
This model allows rich heterogeneity in preferences and price responsiveness across consumers, and it
gains efficiency and precision from simultaneously learning consumer preferences across many
product categories.
Using recent advances in variational Bayesian inference with stochastic gradient descent allows the
model to remain tractable on the types of relatively large data sets that are increasingly becoming
available as digitization progresses.
We show that this approach can yield substantial improvements in out of sample predictive accuracy.
This model is also able to predict price elasticities and patterns of substitution between products,
which are often ignored or explicitly assumed away in most of the related recommender systems
literature from computer science.
Using the nested functional form, inspired by the nested logit model, allows our model to more
efficiently learn these patterns of cross product substitution.
We demonstrate an approach for validating a model's ability to make predictions for counterfactual
questions, by leveraging the large number of price changes and changes in product availability that
occur in the data.
Treating each such change as a ``mini experiment'', we can evaluate a model's predictions before and
after the change on held-out data that was not used to fit the model.
Pooling across many such sources of variation in the data reduces the noise and allows us to compare
models in terms of their ability to make counterfactual predictions.
We evaluate the potential gains from using flexible personalized models such as the one we propose
here for targeting marketing efforts such as personalized price discounts or for identifying
new consumers who might be interested in trying a product.
More generally, we believe that flexible models of consumer demand, such as the Nested Factorization
model proposed here, can be a useful tool for guiding the marketing strategies of firms or as part
of a larger model for understanding patterns of competition between firms.

\section*{Declarations and Acknowledgments}

\begin{itemize}
\item Acknowledgments:
  We are grateful to Tilman Drerup and Ayush Kanodia for exceptional research assistance.
  We thank the seminar participants at Harvard Business School, Stanford, the Microsoft
  Digital Economy Conference, and the Munich Lectures.
\item Funding:
  We acknowledge generous financial support from Microsoft Corporation, the Sloan Foundation, the
  Cyber Initiative at Stanford, and the Office of Naval Research grant N00014-17-1-2131.
  Robert Donnelly is currently employed at Instacart (San Francisco, USA) but
  contributed to this research while a graduate student at Stanford Graduate
  School of Business.
  Ruiz is currently affiliated with DeepMind (London, UK) but contributed to
  this research while at Columbia University (New York, USA) and the
  University of Cambridge (London, UK), supported by the EU H2020 programme
  (Marie Sk\l{}odowska-Curie grant agreement 706760).
  The views expressed herein do not necessarily represent the views of Instacart
  or DeepMind.
\item Availability of data and materials:
  The dataset is available to researchers at Stanford and Berkeley by
  application; it has been used previously in other research papers (see
  https://are.berkeley.edu/SGDC).
\item Code availability: https://www.github.com/rodonn/nested-factorization
  Additional implementations of related variational inference models are available at
  https://github.com/franrruiz/shopper-src and https://www.openicpsr.org/openicpsr/project/114442/version/V1/view
\end{itemize}

\section{Appendix}
\subsection{Data Construction and Sample Selection}\label{sec:AppendixData}

The filters we use to select categories for study are outlined as follows:
\begin{enumerate}
\item For many of the mixed and nested logit specifications, we encountered difficulty with
  convergence in some of the product categories. To reduce these issues we ran all of the logit
  specifications using the top 10 items in each category along with an eleventh ``pooled'' option
  that combined all of the less popular items in the category. The NF and HPF models were run
  without any pooling of items. To make for a fair comparison, we evaluate model fit using only the
  top 10 items in each category. The relative performance of the NF model improves further if we
  compare the sum of the predicted purchase probabilities for the pooled items to the pooled item
  prediction from the logit models.
\item We eliminate categories in which more than 15\% of shopping trips contain multiple items from
  the category or more than 10\% of trips contain multiple top 10 items. Since for these categories
  the assumption of unit demand was substantially violated. For any remaining shopping trips in
  which multiple items from the same category were purchased, we selected one item at random from
  among the purchased items (and treated the remaining items as unpurchased).
\item We eliminate categories where the average absolute within-category correlation of the top 10
  items' prices is greater than $0.75\%$. This address the challenge of
  identifying cross-price elasticities in a handful of categories in which
  virtually all prices move in parallel.
\item We only include categories where at least 2 of the top 10 items have price variation from
  Tuesday to Wednesday in one of the sample weeks and at least 1 of them top 10 UPCs has price
  changes of at least 10 cents in a least 10\% of the sample weeks.
\item We eliminate the top 15\% of categories with the strongest demand seasonality. For each UPC, we first
  calculate seasonality as the Herfindahl index of daily demands over the sample period. We then
  calculate the percentile of each UPCs Herfindahl index over all UPCs and define a category's
  seasonality as the average of the category's top 10 items' percentiles. While
  our approach of category level time controls at the week level should be able
  to control for any category level seasonality, it is not able to control for
  seasonal trends that affect individual UPCs.
\end{enumerate}

The pricing information in our data comes from the transactions, which means we
need to infer the prices a customer would have paid for any items they did not
purchase. In addition, we need to account for coupons and deals (e.g.\ buy 2 get
1 free) that may cause different customers shopping on the same day to pay
different price per unit. To resolve this, we use the daily median transacted price
per unit calculated. In the event of a day with zero purchases, we carry forward
the price data from the previous day.

Our data on product availability (i.e. out-of-stock items) is at granular level
as described \citet{Che2012} based on the times employees scan items as out of
stock and when they are restocked.
However, for simplicity all of the models are run at a daily level.
We consider an item unavailable to all shoppers on any days in which it is listed as out-of-stock
during more than 75\% of shopping trips on that day.

\subsection{Variational Inference Algorithm}\label{sec:AppendixModel}%

In this section, we provide additional details on the implementation of the
variational inference model that is used to estimate the Nested Factorization
model.

Recall from Section~\ref{sec:VB} that for the product choice stage of the model,
we would like to approximate the posterior distribution of the latent parameters
$\ell_p = \{\theta_i,\beta_j,\rho_j,\sigma_i,\gamma_i,\lambda_j\}$.
For notational convenience we will rewrite this as $\ell = \{\ell_1, \ldots,
\ell_K\}$ where the number of latent parameters $K = 3N + 3J$, where $N$ is the
number of households and $J$ is the number of products.
We will approximate the posterior using a multivariate Gaussian distribution
with a diagonal covariance matrix.
In general, imposing this ``mean-field'' assumption may limit the ability
of our variational distribution to approximate the exact posterior, however
this structure often works quite well in practice and still allows
substantial flexibility in the resulting posterior approximation.
\begin{align} \label{eq:variationalformula}
  q\left( \ell ; \nu \right)
  = \mathcal{N}(\ell; \mu, \Sigma)
  = \prod_{k=1}^K \mathcal{N}(\ell_k; \mu_k, \sigma^2_k)
\end{align}
We want to find the variational parameters
$\nu = \{\mu_1, \ldots, \mu_K, \ldots, \sigma^2_K\}$
that minimize the KL divergence between the variational distribution
$q\left( \ell ; \nu \right)$
and the exact posterior
$p(\ell \mid \vect{y}, \vect{x})$.
\begin{equation}
q^*(\ell; \nu) = \argmin_\nu KL\left( q(\ell; \nu) \mid \mid p(\ell \mid \vect{y}, \vect{x}) \right)
\end{equation}

This can be rearranged to show that minimizing the KL
divergence is equivalent to maximizing an expression known as the evidence lower
bound (ELBO) \citep{Blei2017}.
\begin{align}
  \mathscr{L}(\nu)
  &= E_{q(\ell;\nu)} \left[
      \log p(\vect{y} \mid \vect{x}, \ell)
      - \log q(\ell; \nu)
    \right]
  \nonumber \\
  &= E_{q(\ell;\nu)} \left[
      \sum_t \log p(\vect{y}_t \mid x_t, \ell )
      + \log p(\ell)
      + \log q(\ell; \nu)
    \right]
    \label{eq:ELBO_Appendix}
\end{align}

Unfortunately, the expectation in Equation~\ref{eq:ELBO_Appendix} is
analytically intractable.
However, we can still seek the value of $\nu$ that maximizes $\mathscr{L}(\nu)$
with stochastic gradient descent if we are able to find a tractable expression
for an unbiased estimate of the gradient $\nabla_\nu \mathscr{L}(\nu)$.
We can do this by applying an approach that is known as the reparametrization
trick \citep{Kigma2014,Titsias2014,Rezende2014}.

To do this, we introduce a transformation of the latent variables,
so that rather than directly drawing from the distribution
$ \ell \sim q\left( \ell ; \nu \right)$, we instead draw a new auxiliary random
variable $\varepsilon \sim \mathcal{N}(0, I_K)$ from a standard
multivariate Gaussian distribution.
By applying the transformation
$\mathcal{T}(\varepsilon; \nu) = \mu + \Sigma^{\frac{1}{2}} \varepsilon$
we can generate draws $\ell = \mathcal{T}(\varepsilon; \nu)$ such that
$\ell \sim q(\ell; \nu)$.

For notational ease, we will denote the expression inside the expectation in
Equation~\ref{eq:ELBO_Appendix} as
$f(\ell; \nu)$.
We can now rewrite the expectation in the gradient of the ELBO as
\begin{equation}
  \nabla_\nu \mathscr{L}(\nu)
  = \nabla_\nu E_{q(\ell;\nu)} \left[ f(\ell; \nu) \right]
  = \nabla_\nu E_{\varepsilon} \left[ f(\mathcal{T}(\varepsilon; \nu); \nu) \right]
\end{equation}

Now, bringing the gradient inside of the expectation and applying the chain rule
gives us
\begin{equation}
  \nabla_\nu \mathscr{L}(\nu)
  = E_{\varepsilon} \left[
    \nabla_\ell f(\ell; \nu); \nu)\mid_{\ell = \mathcal{T}(\varepsilon; \nu)}
    \nabla_\nu \mathcal{T}(\varepsilon; \nu)
  \right]
\end{equation}
To obtain this expression, we used the fact that
$E_\varepsilon \left[
  \nabla_\nu f(\ell, \nu)\mid_{\ell = \mathcal{T}(\varepsilon; \nu)}
\right] = 0$, since the only dependence of $f(\ell; \nu)$ on $\nu$ is through
the term $\log q(\ell; \nu)$ and the expected value of the score function is 0.

Now we can obtain a Monte Carlo estimate of the gradient of the ELBO by sampling values
of $\varepsilon$ from the standard multivariate Gaussian in order to approximate
this expectation.
In addition to making draws of $\varepsilon$ to evaluate the expectation, we
also subsample customer shopping trips in each iteration.
This allows the estimation to more easily scale to large datasets.
By scaling the gradient estimates to account for this sampling, we can maintain
the unbiasedness of the gradient estimator.

\subsubsection{Hyperparameters}
For each of the variational parameters we use a prior variance of 1.0 and a
prior mean of 0 and a batch size of 5000 (number of customer trips to sample in
each step of the stochastic gradient descent).
We do a hyperparameter grid search across:
\begin{itemize}
\item \{20,40,80\} for the dimension of the latent factorization $\theta_i \beta_j$
\item \{20,40,80\} for the dimension of the price term $\gamma_i \lambda_j$
\item With and without user demographics $W_{i}$
\item With and without product characteristics $X_j$
\item \{0.001,0.005,0.01\} learning rates (i.e. step size for gradient descent)
\item Product level model only: whether price should enter linearly or in logs
\item Category level model only: \{10,20\} for the dimension of the time factorization $\mu_{c} \delta_t$
\end{itemize}

We selected models based on counterfactual price performance on the validation
set.
The selected product level model has hyperparameters \{80, 20, yes, no, 0.005, linear price\}.
The selected category level model has hyperparameters \{40, 40, no, no, 0.01, 10\}.

\subsection{Estimated Elasticities and Purchase Probabilities}

In this section we compare the predicted own price elasticities across models
and how those predicted elasticities vary within and between products.
Table~\ref{tab:ElasticityOwn} shows the median own price elasticity estimated
by each model.
SD(Mean) is the standard deviation across the mean product level own-price
elasticities. This captures how much variability there is in elasticities
across products.
Mean(SD) is the mean of the standard deviation of
elasticities across consumers within a specific product. This captures the amount
of variability in elasticities across consumers for the same product.

\begin{figure}
  \centering
  \includegraphics[width=\fullw]{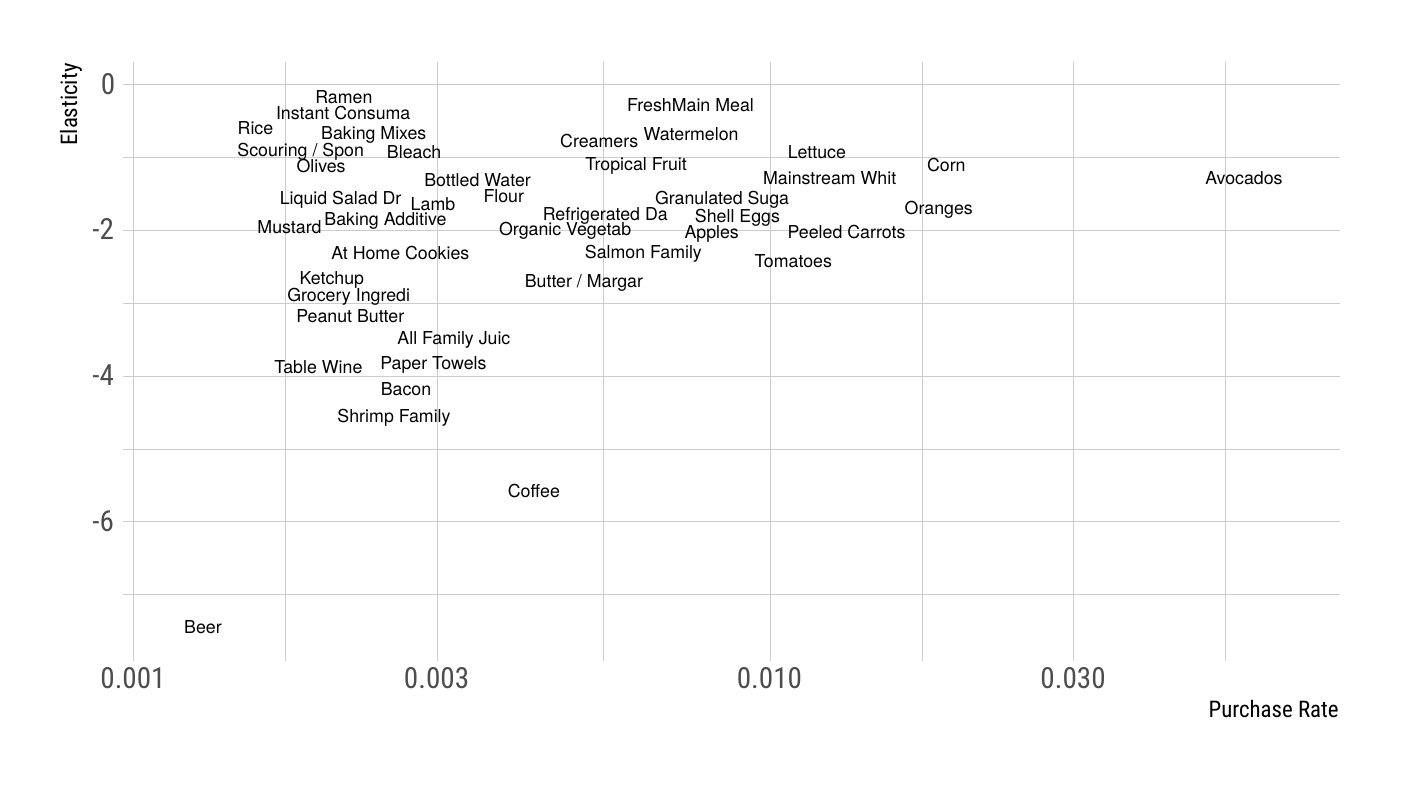}
  \caption{Category Level Elasticities and Predicted Purchase Probabilities:\\
  }\label{fig:CategoryElasticityPropensity}
\end{figure}

\begin{figure}
  \centering
  \includegraphics[width=\fullw]{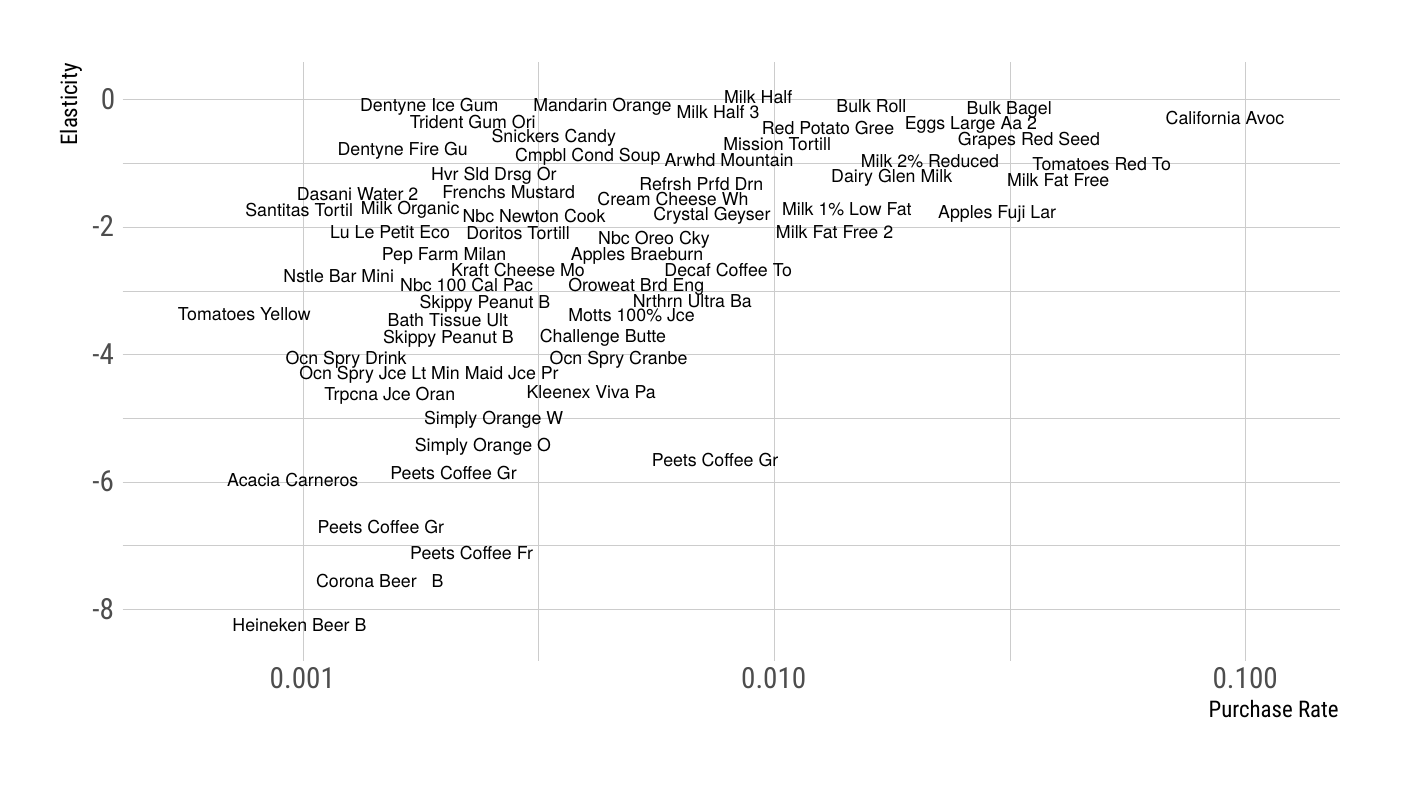}
  \caption{UPC Level Elasticities and Predicted Purchase Probabilities:\\
  }\label{fig:UpcElasticityPropensity}
\end{figure}

Figures~\ref{fig:UpcElasticityPropensity} and
\ref{fig:CategoryElasticityPropensity} show the predicted elasticities and
purchase rates for a sample of products and categories (entries
are hidden when their text box would overlap with another entry).

\begin{table}
  \caption{Comparison of Own-Price Elasticities}\label{tab:ElasticityOwn}
  \centering
  \resizebox{\linewidth}{!}{
    \begin{tabular}{lccccccccc}
      Model                                               & Median  & SD(Mean) & Mean(SD) \\
      \midrule
      Nested Factorization                                & -1.7121 & 1.2008   & 1.7774 \\
      Nested Logit with Demographic Controls              & -1.2976 & 1.0377   & 0.0532 \\
      Mixed Logit with Random Price and Random Intercepts & -2.2024 & 1.7822   & 0.8387 \\
      Mixed Logit with Random Price and HPF Controls      & -2.7077 & 1.9084   & 1.1294 \\
      Multinomial Logit with HPF Controls                 & -1.1841 & 0.8904   & 0.0060 \\
      Multinomial Logit with Demographic Controls         & -1.1813 & 0.8837   & 0.0017 \\
      Mixed Logit with Random Price and Demographics      & -3.0897 & 2.4546   & 1.4785 \\
      Nested Logit with HPF Controls                      & -1.1017 & 0.8783   & 0.0182 \\
      \bottomrule
      \bottomrule
      \addlinespace
    \end{tabular}}
\end{table}

\end{document}